\documentclass[review]{elsarticle}

\usepackage{lineno,hyperref}
\modulolinenumbers[5]

\journal{Elsevier}

\usepackage[english]{babel}
\usepackage[utf8]{inputenc}

\usepackage{url}
\usepackage{makeidx}         % allows index generation
\usepackage{multicol}        % used for the two-column index
\usepackage[bottom]{footmisc}% places footnotes at page bottom
\let\proof\relax 

\usepackage{amsmath,bm,amsfonts,amssymb}
\usepackage{amsthm}

\usepackage{commath}      % for norm 

\usepackage{xpatch}

\usepackage{color,xcolor,ucs}
\usepackage{subfig}
\usepackage[font=small,labelfont=bf]{caption}
\usepackage{tabularx}
\usepackage{float}
\usepackage{wrapfig}

\usepackage{algorithm}
\usepackage[noend]{algpseudocode}% http://ctan.org/pkg/algorithmicx
\usepackage{multirow}
\usepackage{listings}
\usepackage{lipsum}
\usepackage{textcomp} % for registered symbol

\usepackage[nocomma]{optidef}  % for optimization objective function definition

\usepackage{graphicx}
\usepackage{comment}
\usepackage[export]{adjustbox}

\usepackage{siunitx} %for table number alignment

\usepackage{colortbl}  % table color change
\makeatletter
\algnewcommand{\LineComment}[1]{\Statex \hskip\ALG@thistlm \(\triangleright\) #1}
\algnewcommand{\LineCommentCont}[1]{\Statex \hskip\ALG@thistlm \parbox[t]{\linegoal}{\hangindent=1em\hangafter=1 $\triangleright$ #1}}
\makeatother

\theoremstyle{definition}

\newtheorem{theorem}{Theorem}
\xpatchcmd{\proof}{\itshape}{\normalfont\proofnamefont}{}{}

\newcommand{\proofnamefont}{\bfseries}
\newcommand{\C}{\mathcal}
\definecolor{mygray}{gray}{0.6}

\definecolor{revisioncolor}{rgb}{0.1,0.1,1}
\newcommand{\revtext}[1]{{\color{revisioncolor} #1}}

\definecolor{mybrown}{rgb}{0.314,0.086,0.086}

\begin{document}

\begin{frontmatter}

\title{An Incremental Sampling and Segmentation-Based Approach for Motion Planning Infeasibility}

\author[label1]{Antony Thomas}
\ead{antony.thomas@iiit.ac.in}
\address[label1]{Robotics Research Center, IIIT Hyderabad, Hyderabad 500032, India.}

\author[label2]{Fulvio Mastrogiovanni}
\ead{fulvio.mastrogiovanni@unige.it}
\address[label2]{Department of Informatics, Bioengineering, Robotics, and Systems Engineering, University of Genoa, Via All'Opera Pia 13, 16145 Genoa, Italy. }

\author[label2]{Marco Baglietto}
\ead{marco.baglietto@unige.it}
           
\begin{abstract}
We present a simple and easy-to-implement algorithm to detect plan infeasibility in kinematic motion planning. Our method involves approximating the robot's configuration space to a discrete space, where each degree of freedom has a finite set of values. The obstacle region separates the free configuration space into different connected regions. For a path to exist between the start and goal configurations, they must lie in the same connected region of the free space. Thus, to ascertain plan infeasibility, we merely need to sample adequate points from the obstacle region that isolate start and goal. Accordingly, we progressively construct the configuration space (initially assumed to be entirely free) by sampling from the discretized space and updating the bitmap cells representing obstacle regions. Subsequently, we partition this partially built configuration space to identify different connected components within it and assess the connectivity of the start and goal cells. We illustrate this methodology on five different scenarios with configuration spaces having up to 5 degrees-of-freedom (DOF). Additionally, we discuss further optimizations designed to significantly accelerate the proposed algorithm. The scalability of our approach to higher-dimensional configuration spaces is also examined, with experimental demonstrations involving 6-DOF and 7-DOF robots.
\end{abstract}


\end{frontmatter}
\section{Introduction}
Motion planning is a fundamental problem in robotics, involving finding a path for a robot from its start configuration to a goal configuration without colliding with obstacles. A complete motion planner can either compute a collision-free path from the start to the goal or conclude that no such path exists. However, complete motion planning is challenging, and most approaches focus on finding a feasible plan with weaker notions of completeness. Resolution complete planners, typically those based on cell decomposition, offer completeness provided that the number of cells used to discretize the configuration space is sufficiently high~\cite{zhang2007IROS}. Yet, in high-dimensional configuration spaces, such approaches tend to be computationally very expensive. Sampling-based motion planners~\cite{kavraki1996IEEE, kuffner2000ICRA} are typically employed in such cases to find paths as quickly as possible. However, they are only probabilistically complete~\cite{karaman2011IJRR}, meaning that if a plan exists, they will find it given enough time, but if no plan exists, they can run forever (or until a timeout). Therefore, a timeout is not a guarantee of infeasibility. In this work, we focus on the less examined path non-existence problem and present a simple algorithm that checks for motion planning infeasibility.
\begin{figure}[t!]
\centering
  \subfloat[]{\includegraphics[scale=0.55]{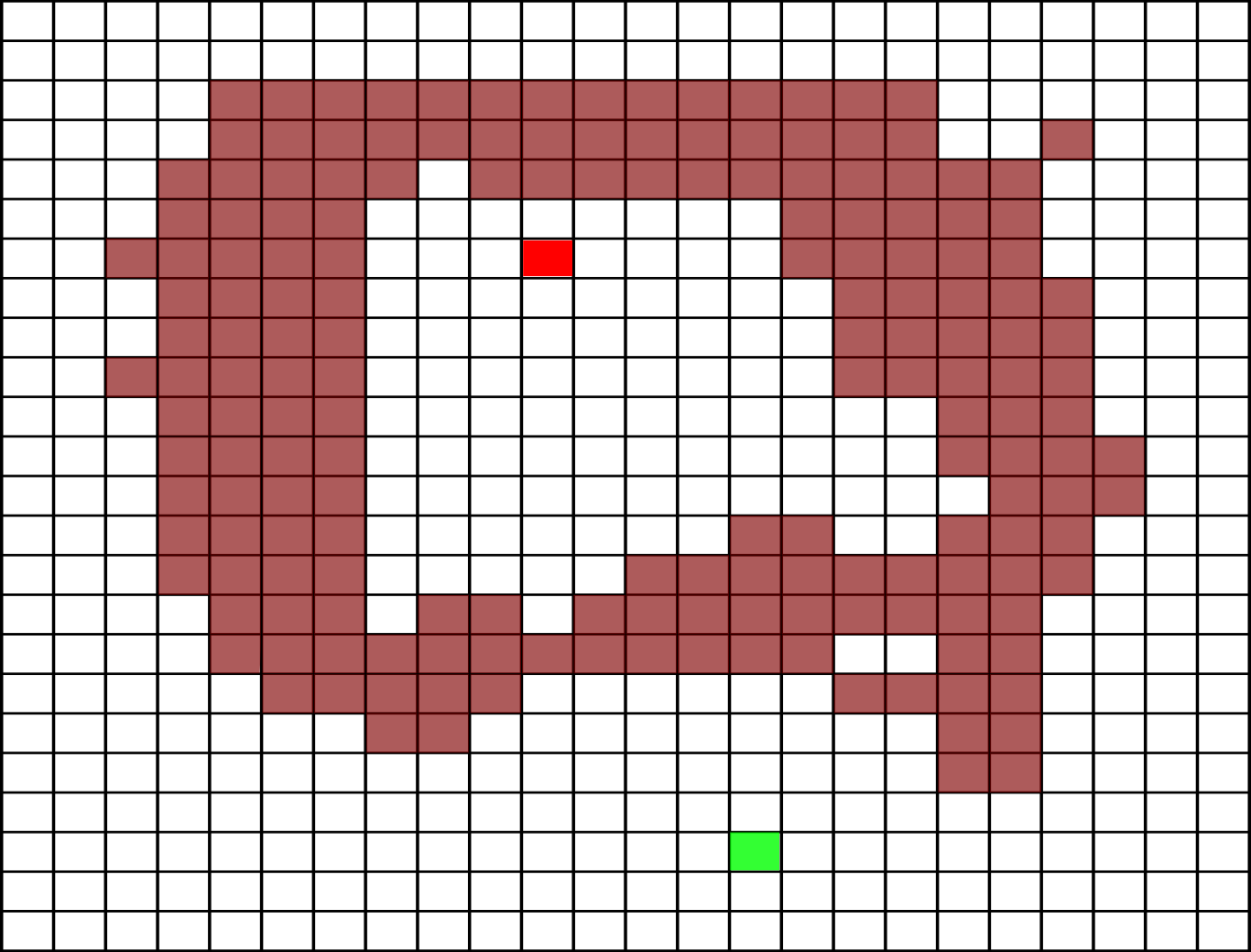}\label{fig:concept1}}\hspace{0.1cm}
      \subfloat[]{\includegraphics[scale=0.55]{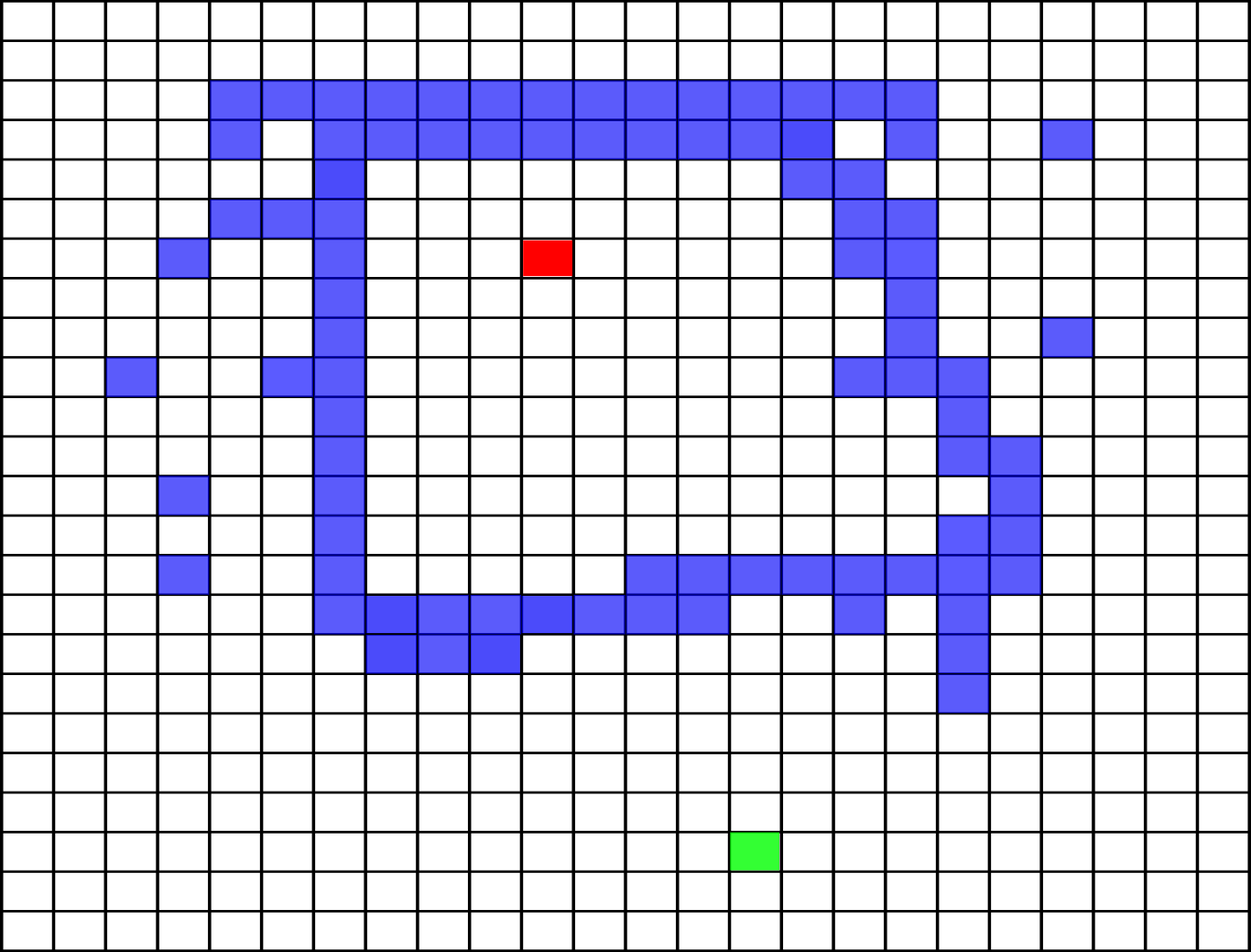}\label{fig:concept2}}\\
             \subfloat[]{\includegraphics[scale=0.55]{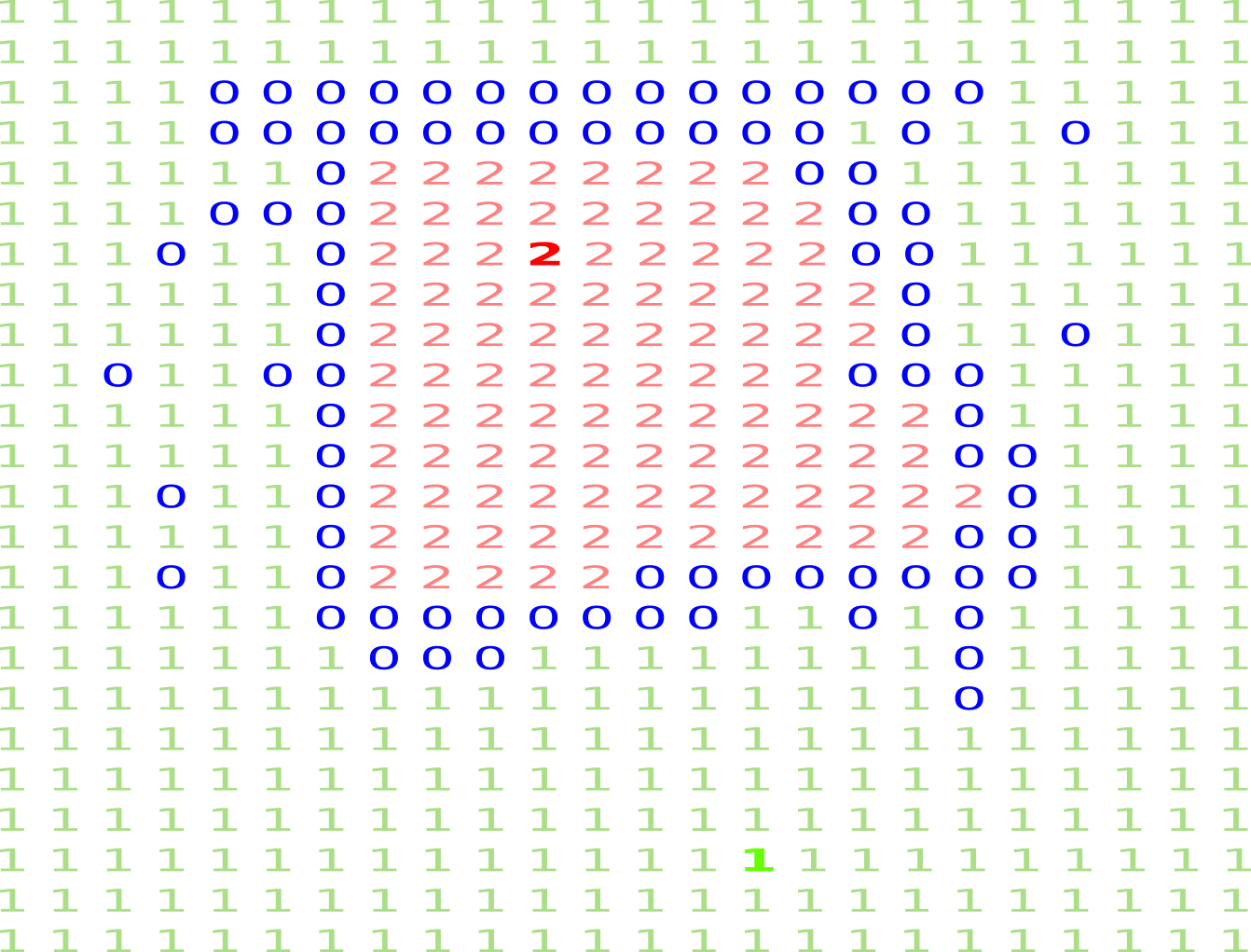}\label{fig:concept3}}
          \caption{(a) A representative discrete configuration space with the start and goal configuration cells colored in green and red, respectively. Other colored cells represent obstacle regions, while uncolored cells represent free space. (b) The sampled obstacle region is colored in blue. This partially constructed configuration space is sufficient to establish motion planning infeasibility since there exists no path from the start to the goal. (c) Segmented representation of the configuration space shown in (b). There are two regions denoted by 1's and 2's, respectively, separated by the obstacle region indicated by 0's. Since the start configuration belongs to region 1 and the goal configuration belongs to region 2, motion planning is infeasible.}
  \label{fig:concept}
\end{figure}

Motion infeasibility is a critical aspect of many robot planning methodologies. Task and motion planning~\cite{kaelbling2013IJRR, srivastava2014ICRA, lagriffoul2014IJRR, dantam2018IJRR, garrett2018IJRR, thomas2021RAS} must consider the feasibility of motion plans to achieve the associated high-level tasks. When motion planning is deemed infeasible, alternative task plans must be generated. Similarly, feasibility checks are fundamental in manipulation tasks amidst clutter or rearrangement planning~\cite{stilman2007ICRA, dogar2011RSS, krontiris2015RSS, karami2021AIIA}. This often entails either displacing obstacles obstructing the task, usually identified through motion planning infeasibility, or positioning them at specific locations. The latter requires evaluating the feasibility of motion plans for different object placements. Similarly, in Navigation Among Movable Obstacles (NAMO)~\cite{stilman2005IJHR, muguiraIturralde2023ICRA}, the environment is actively modified by rearranging obstacles to create feasible paths.

\textit{The main contribution of this paper is a simple and easy-to-implement algorithm for proving the infeasibility of motion planning}.\footnote{In Section~\ref{subsec:feasible_infeasible}, we show that the approach can be naturally extended to incorporate plan feasibility by running a motion planner in parallel with the proposed method. In this setting, a feasible plan is returned if one exists; otherwise, the proposed method provides an infeasibility certificate.} Our framework is based on discretized configuration spaces which preserve the essential structure of their continuous counterparts. We refer to these as \textit{equivalent} spaces, discussed in detail in Section~\ref{sec:approach}. The key insight of our approach is that proving infeasibility does not require constructing the entire configuration space obstacle region--- only a relevant subset (see Fig.~\ref{fig:concept}).

We introduce a technique to demonstrate motion planning infeasibility by segmenting a discrete configuration space into distinct free regions isolated by the obstacle region. We note that such distinct free regions are referred to as connected components of the free space. Therefore, we use the terms \textit{distinct free region} and \textit{connected component} interchangeably. If the start and goal configurations are partitioned into separate free regions, motion planning is deemed infeasible. 

However, computing a discrete configuration space is computationally expensive, especially for a robot with high degrees-of-freedom (DOF). To address this issue, we initially set the entire configuration space to free regions. Subsequently, we incrementally construct the configuration space by sampling and checking if the sampled configurations lie within the obstacle region. The key concept underpinning the incremental sampling is based on the principle that it is often unnecessary to sample the entire obstacle region. Instead, our objective is to draw a sufficient number of samples that establish the partition of the start and goal configurations into disconnected components of the configuration space. During each iteration, the partially constructed configuration space is segmented into distinct connected free regions as shown in Fig.~\ref{fig:concept3}. The segmented configuration space is then queried to determine whether the start and the goal configurations belong to the same connected free region. In Section~\ref{sec:approach}, we elucidate this approach in detail and further discuss methods to speed up the construction of the configuration space.

In Sections~\ref{sec:rel1} to~\ref{sec:rel3}, we provide an overview of previous works related to motion planning infeasibility. These works are classified based on the different strategies employed to establish the non-existence of a feasible path. In Section~\ref{sec:compare}, we highlight how our approach differs from these methods in terms of computational cost and scalability.
\subsection{Approximate Configuration Space} 
\label{sec:rel1}
Zhang~\textit{et al.}~\cite{zhang2008IJRR} decompose the configuration space into cells, which are then queried to determine if they lie within the obstacle region. Subsequently, a graph is constructed where the nodes represent the cells and the edges represent adjacent cells. Using the cells that contain free regions, the problem of path non-existence is transformed into a graph search problem. Checking the occupancy of each cell is computationally challenging in higher dimensions. In contrast, our approach does not always require checking the occupancy of every cell. In a different approach presented in~\cite{mccarthy2012ICRA}, obstacle regions are decomposed into collections of simplices called alpha shapes. These simplices are then utilized to address connectivity queries. However, methods for computing higher-dimensional alpha shapes are presently unknown. Points in the obstacle region are sampled to identify possible facets of a separating polytope in \cite{li2020IROS}. A separating polytope is a closed polytope that separates the start and goal into disconnected components of the free configuration space. However, the generation of these facets is computationally expensive, which can lead to scalability issues, especially in higher-dimensional configuration spaces. Varava~\textit{et al.}~\cite{varava2021IJRR} construct an
approximation of the obstacle region by decomposing it into a set of slices corresponding to subspaces of fixed obstacle orientations. They then compute the free space as the complement of this approximated obstacle region, which is subsequently used to synthesize a connectivity graph. A rigid body passing through a narrow gate is considered in~\cite{basch2001ICRA}. The orientations of the rigid body are discretized, and each orientation is individually checked for its ability to pass through the gate. 
\subsection{Learning-Based Infeasibility Estimation}
\label{sec:rel2}
 In~\cite{sung2023RSS}, a representative roadmap is learned from available training problems, along with the probability of the edges being collision-free. Yet, the authors only prove the infeasibility in the roadmap and not the configuration space. Approaches in~\cite{wells2019RAL,driess2021IJRR} learn a classifier that guides the robot toward feasible motions. However, the classifier is used only as a heuristic to quickly estimate the feasibility of high-level actions rather than as an infeasibility proof. Li~\textit{et al.}\cite{li2023IJRR} combine supervised learning and sampling-based planning to certify motion planning infeasibility. Their approach constructs an infeasibility proof by learning a manifold that lies entirely within the obstacle region and querying whether it separates the start and goal configurations. However, their method assumes that it is always possible to sample points on the manifold and compute the configuration-space penetration depth of these samples. This approach is demonstrated for robots with up to 4 degrees of freedom (DOF). Scalability to 5-DOF robots is addressed in\cite{li2023RAL}, where the triangulation method used to represent the learned manifold in~\cite{li2023IJRR} is significantly improved.
\subsection{Feasibility Through Constraint Modification} 
\label{sec:rel3}
The minimum constraint displacement (MCD)~\cite{hauser2013RSS, thomas2022IAS, thomas2023ICRA} and the minimum constraint removal (MCR)~\cite{hauser2014IJRR, thomas2023IAS} motion planning problems identify the minimum displacement of obstacles and the minimum number of obstacles to be removed, respectively, to guarantee a feasible motion plan. MCD class of problems does not inherently provide a means to prove infeasibility. If planning is infeasible, MCD computes the minimum displacement of obstacles required to ensure feasibility. The MCR class of problems is generally solved by partitioning the configuration space along the obstacle boundaries to obtain different connected regions that form a discrete graph. The MCR graph is then queried to determine the minimum number of obstacles to be removed to connect the start and the goal configuration. Implicitly, if the start and goal configurations are disconnected by obstacle regions, it implies that motion planning is infeasible. However, computing such partitions becomes intractable as the number of obstacles increases, especially in high-dimensional spaces.
\subsection{Comparative Analysis with Prior Approaches}
\label{sec:compare}
For most approaches categorized as \textit{approximate configuration space} or \textit{learning-based}, the input typically consists of an approximate configuration space computed using techniques such as cell decomposition or sampling-based methods~\cite{kavraki1996IEEE,kuffner2000ICRA}. Importantly, the overall computation time for determining infeasibility in these methods does not account for the time required to construct the approximate configuration space, as it is provided as input. In contrast, our approach incrementally constructs the configuration space during execution, which constitutes the primary computational effort, as demonstrated in Section~\ref{sec:experiments}. 

MCD and MCR implicitly address infeasibility by identifying minimal obstacle displacements or removals that restore feasibility. However, these problems are NP-hard and scale poorly with increasing dimensionality.

Current state-of-the-art infeasibility certification methods scale only up to 4- or 5-DOF robots~\cite{li2023IJRR,li2023RAL}, with experiments completing in minutes using parallel computing. While our experiments use different benchmark scenes, we achieve comparable runtimes for 4- and 5-DOF robots, even in the absence of parallel computing. Moreover, by leveraging quotient-space topology~\cite{orthey2018IROS}, we extend our approach to 6-DOF and 7-DOF manipulators, thereby demonstrating its practical feasibility for high-dimensional motion planning infeasibility certification.

%
%\section{Related Work and Comparison}
%\input{related}
%
\section{Preliminaries}
\label{sec:preliminaries}
We consider a robot $\C{R}$, which operates in the workspace $\C{W} \subset \mathbb{R}^a$, with $a = 2$ or $a = 3$. An obstacle in $\C{W}$ will be denoted by $\C{O}_i \in \C{O}$ and $\C{C}$ will be used to denote the configuration space (C-space in short) of the robot $\C{R}$. A configuration $q\in \C{C}$ of $\C{R}$ completely specifies the volume occupied by $\C{R}$, and will be denoted by a list of $n$ parameters $q = (q_1,\ldots,q_n)$, where $n$ is the dimension of $\C{C}$. Therefore, given a configuration $q$ of $\C{R}$, the corresponding placement of $\C{R}$ in $\C{W}$ will be denoted by $\C{R}(q)$. Every obstacle $\C{O}_i \subset \C{W}$ maps into $\C{C}$ to a C-obstacle $=\{q\in \C{C}\ | \ \C{R}(q) \cap \C{O}_i \neq \emptyset \}$ and the union of all C-obstacles is the C-obstacle region. Finally, C-free is the region in $\C{C}$ given by $\C{C}\setminus$C-obstacle region, that is, C-free $=\{q\in \C{C}\ | \ \C{R}(q) \cap \C{O} = \emptyset\}$. The start and goal configurations will be denoted as $q_s$ and $q_g$, respectively.

In this work, we limit the C-space to a discrete set, where each DOF is constrained to finite values. This discretized C-space will be called the C-space bitmap $\C{CB}$. Thus, for a C-space of dimension $n$ (corresponding to $n$ DOF), we obtain an $N_1 \times N_2 \times \ldots \times N_n$ C-space bitmap, denoted by $\C{CB}$, where $N_i$ represents the number of discrete divisions along the $i$-th DOF (or dimension) of the C-space. In other words, $\C{CB}$ is a grid cell represented as a multidimensional array of size $N_1 \times N_2 \times \ldots \times N_n$, and the resolution along the $i$-th dimension is given by $(\overline{q}_i - \underline{q}_i)/N_i$, where $\underline{q}_i$ and $\overline{q}_i$ denote the lower and upper joint limits, respectively. Each cell of the $n$-dimensional $\C{CB}$ corresponds either to the C-free (denoted by 1) or the C-obstacle (denoted by 0)
%--- note that this differs from the more common convention where 0 cells indicate free space. 

As argued above, any configuration $q \in \C{C}$ of $\C{R}$ can be mapped to the subset of $\C{W}$ occupied by $\C{R}$, that is, $\C{R}(q)$. Therefore, $\C{CB}$ may be constructed by iterating over all possible $n$-tuples $q = (q_1,\ldots,q_n)$ belonging to the $N_1\times N_2\times \ldots \times N_n$ binary array such that $\C{CB}(q) = 1$ when $\C{R}(q)$ does not collide with any obstacle and $\C{CB}(q) = 0$, when $\C{R}(q)$ collides with  at least one obstacle. Checking for collision with obstacles for all configurations is computationally expensive, especially when dealing with higher resolutions and dimensions of configuration spaces. For the 4-DOF robot depicted in Fig.~\ref{fig:4dof}, generating the complete $\C{CB}$ with a resolution of $36\times 36\times 36\times 36$ required approximately 58 minutes\footnote{To check for collisions with obstacles, both the robot and obstacle polygons are subdivided into triangles, and a triangle intersection algorithm~\cite{mccoid2022TMS} is used. Computation performed on an Intel{\small\textregistered} Core i7-10510U CPU$@$1.80GHz$\times$8 with 16GB RAM under Ubuntu 18.04 LTS.}. While methods for efficiently computing discrete C-space exist~\cite{kavraki1995TRO, curto1997CIRA}, they tend to become prohibitively expensive as the dimension of the C-space increases.
\section{Approach}
\label{sec:approach}
In a general perspective, our approach incrementally constructs the C-space bitmap $\C{CB}$, by sampling configurations in C-obstacle. During each iteration, the updated $\C{CB}$ is segmented into different sets of adjacent \textit{free} regions, akin to identifying distinct groups or regions of connected pixels in an image. We subsequently check whether the start ($q_s$) and the goal ($q_g$) reside in separate regions. Motion planning is infeasible if $q_s$ and $q_g$ are in different C-free regions. If the check fails, the iteration is repeated.
\begin{algorithm}[t]
\caption{Motion Planning Infeasibility Detection}
\label{algo:MPI}
\begin{algorithmic}[1]
\Require{$\C{R}, q_s, q_g, \C{O}, n, ns, d$}
\LineComment \textcolor{mygray}{$ns$ denotes the minimum number of C-obstacle configurations sampled in each iteration.}
\State{$\C{CB} \gets true(N^1,\ldots,N^n)$}
\State{$SS\gets true(sc)$}
\LineComment \textcolor{mygray}{Array with $sc= N^1* N^2* \ldots * N^n$, the size of $\C{CB}$.}
\While{true}
\State{$\C{CB}, SS \gets$ $\texttt{SampleCobstacle} \left(\C{R}, \C{O}, SS, ns, d\right)$}
\State{$L \gets \texttt{SegmentCB} \left( \C{CB}\right)$}
\If{$\texttt{SegmentCheck}(L,q_s,q_g)$}
\State \Return{\text{No Motion Plan}}
\EndIf
\EndWhile
% \State \Return{\text{Motion Plan Exists}}
\end{algorithmic}
\end{algorithm}

The overall approach is summarized in Algorithm~\ref{algo:MPI}. Initially, all cells in the bitmap $\C{CB}$ are initialized to 1 (C-free). In each iteration, the subroutine \texttt{SampleCobstacle} is invoked (Algorithm~\ref{algo:MPI}, line 4), which samples at least $ns$ configurations lying in C-obstacle. The cells of $\C{CB}$ corresponding to these C-obstacle samples are then set to zeros. The updated $\C{CB}$ is segmented (line 5) to check the connected components to which $q_s$ and $q_g$ belong (line 6). If they are in different connected components, motion planning is infeasible. However, if they are in the same connected component, either a motion plan exists, or $\C{CB}$ must be further updated with additional samples from the C-obstacle to establish infeasibility, continuing the iteration to draw new samples as needed.
\subsection{Sampling C-obstacle}
\begin{algorithm}[]
\caption{\texttt{SampleCobstacle}}
\label{algo:samplecobstacle}
\begin{algorithmic}[1]
\Require{$\C{R}, \C{O}, SS, ns, d $}
\For{$i = 1,2,\ldots,ns$}
\While{true}
\State{$ idx \gets \texttt{RandomSample}(SS)$}
\State{$SS(idx) \gets false$}
\State{$q \gets \texttt{GetConfig}(\C{CB}[idx])$}
\LineComment \textcolor{mygray}{Returns $q$ corresponding to $\C{CB} $ cell index $idx$.}
\State{$cc, p \gets \texttt{CollisionCheck}(\C{R}(q), \C{O})$}
\LineComment \textcolor{mygray}{$p$ is the set of links of $\C{R}$ that are in collision with obstacles for the configuration $q$.}
\If{$cc$}
\State{$\C{CB}(idx) \gets false$} 
\State{$\C{CB}, SS \gets \texttt{SpeedUp}(q,p)$}
\For{$d$ neighbors of $q$}
\LineComment \textcolor{mygray}{$d$ neighbors of $q$ randomly selected.}
\State{$SS(idx_d)\gets false$}
\LineComment \textcolor{mygray}{$idx_d$ is the index of neighbor $d$.}
\State{$q \gets \texttt{GetConfig}(\C{CB}[idx_d])$}
\State{$cc, p \gets \texttt{CollisionCheck}(\C{R}(q),\C{O})$}
\If{$cc$}
\State{$\C{CB}(idx_d) \gets false$}
\State{$\C{CB}, SS \gets \texttt{SpeedUp}(q,p)$}
\EndIf
\EndFor
\State{\textbf{break}}
\LineComment \textcolor{mygray}{Exit the while loop.}
\Else
\State{continue}
\EndIf
\LineComment \textcolor{mygray}{Continue the while loop.}
\If{Timeout}
\State{\Return{$\C{CB}, SS$}}
\EndIf
\EndWhile
\EndFor
\State{\Return{$\C{CB}, SS$}}
\end{algorithmic}
\end{algorithm}
The fundamental concept underlying incremental sampling from the C-space is that it is not necessary to sample the entire C-obstacle region. Instead, it suffices to draw enough samples from the C-obstacle to separate the start and goal configurations into disconnected components within the C-space.

Since we initialize all cells in $\C{CB}$ as C-free, we need to sample configurations in C-obstacle to progressively build $\C{CB}$. We maintain a set $SS$ (Algorithm~\ref{algo:MPI}, line 2) that stores the linear indices $(idx)$ corresponding to the multidimensional array representation of $\C{CB}$. The value $\C{CB}[idx]$ denotes the grid cell associated with index $idx$. Algorithm~\ref{algo:samplecobstacle} samples without replacement from $SS$ (lines 3–4) until at least $ns$ configurations within the C-obstacle are selected. Each sampled configuration $q$ is then checked for collision (Algorithm~\ref{algo:samplecobstacle}, line 6). If $\revtext{q} \in \text{C-obstacle}$, the subroutine \texttt{CollisionCheck} returns \texttt{true} ($cc$), and the corresponding $\C{CB}$ cell is updated to $false$ (Algorithm~\ref{algo:samplecobstacle}, line 8), indicating that the cell belongs to the C-obstacle.

The \texttt{CollisionCheck} subroutine also returns $p$, the set of links of robot $\C{R}$ that are in collision with obstacles for the given configuration $q$. This set $p$ is subsequently used by the \texttt{SpeedUp} subroutine, which enables classification of certain cells as belonging to the obstacle region without explicitly invoking the collision checker. Further details on this procedure are provided in the following paragraphs.

We also check for neighboring configurations that may be in collision (lines 10–16). This leverages the fact that if a configuration is in collision, nearby configurations are also likely to be in collision. To this end, we randomly select $d$ neighbors of each sampled $q \in$ C-obstacle. The total number of possible neighbors of $q$ depends on the C-space dimension, as further discussed in Section~\ref{sec:discussion}. The subroutine \texttt{SampleCobstacle} thus returns an updated $\C{CB}$, which is then segmented into connected components to assess motion-planning infeasibility.

We note that, in each iteration, although $ns$ configurations are explicitly sampled from the C-obstacle, both the \texttt{SpeedUp} subroutine and the selection of $d$ neighboring configurations contribute additional samples within the C-obstacle. Consequently, Algorithm~\ref{algo:samplecobstacle} ensures that at least $ns$ configurations are processed per iteration.

\textbf{Subroutine \texttt{SpeedUp}}: A significant computational bottleneck in Algorithm~\ref{algo:samplecobstacle} is the \texttt{CollisionCheck} subroutine, which, for each sampled configuration $q$, checks for collisions between $\C{R}(q)$ and all obstacles in the environment. To optimize this process, we employ a simple heuristic that allows for avoiding unnecessary collision checks for certain configurations $q$. We will clarify this with an illustrative example. Let us consider an articulated robot with $n$ links. Suppose that the $j$-th link, for a sampled configuration $q = (q_1, \ldots, q_j, \ldots, q_n)$, collides with an obstacle. In such a case, it is readily observed that
\begin{equation}
\forall q \ | \ q = (q_1, \ldots, q_j, \ast) \implies q \in \text{C-obstacle}
\end{equation}
\noindent where $\ast$ denotes the fact that the components $(q_{j+1}, \ldots, q_n)$ may assume any admissible values.  
In other words, for all configurations where the components from the first link to the $j$-th link are fixed (values corresponding to those that resulted in collision with the obstacle), any combinations of the $(j+1)$-th to $n$-th links will still collide with that obstacle. Thus, without querying the \texttt{CollisionCheck} subroutine, all the cells in $\C{CB}$ corresponding to such configurations may be updated. Algorithm~\ref{algo:speedup} summarizes this procedure.
\begin{algorithm}[t]
% \color{revisioncolor}
\caption{\texttt{SpeedUp}}
\label{algo:speedup}
\begin{algorithmic}[1]
\Require{$q = (q_1, \ldots, q_n), p$}
\LineComment \textcolor{mygray}{$p$ is the set of links of $\C{R}$ that are in collision with obstacles for the configuration $q$.}
\State{$j = \min p $}
\ForAll{$\tilde{q} \in \C{CB} \; \text{\textbf{such that}} \;
 \tilde{q}_{1:j} = (q_1, \ldots,  q_j)$}
\State{$\C{CB}(\tilde{q}) \gets false$}
\State{$SS(\tilde{q}) \gets false$} 
\EndFor
\State{\Return{$\C{CB}, SS$}}
\end{algorithmic}
\end{algorithm}

A similar principle has recently been exploited in motion planning through the use of quotient-space decompositions, where infeasible equivalence classes of configurations in the C-space are pruned based on admissibility in lower-dimensional projections~\cite{orthey2019ISRR}. This strategy effectively reduces the sampling space and has been shown to achieve runtime improvements of at least an order of magnitude.

An analogous heuristic can also be applied when the C-obstacle arises due to self-collision. If the $i$-th link and the $j$-th link collide, it is verified that
\begin{equation}
\forall q \ | \ q = (\ast, q_i, \ldots, q_j, \ast) \implies q \in \text{C-obstacle}
\end{equation}
\noindent where $\ast$ denotes the fact that the components $(q_{1}, \ldots, q_{i-1})$, $(q_{j+1}, \ldots, q_n)$ may assume any admissible values.

The process described above is achieved by calling the \texttt{SpeedUp} subroutine (line 9), which takes the colliding link(s) $p$ as input. Similarly, in the case of a mobile robot that can rotate and translate, if a sampled configuration $q$ is such that $q \in \text{C-obstacle}$, and the rotation axis of the robot intersects with an obstacle, then all rotations of the robot at that location will result in a collision. Hence, the configurations corresponding to all possible rotations with the fixed robot location are readily classified as C-obstacle, eliminating the need to query the \texttt{CollisionCheck} subroutine.
\subsection{$\C{CB}$ Segmentation}
Since we only consider kinematic motion planning, infeasibility arises from the obstacle region (C-obstacle) in the C-space. Therefore, a motion plan exists only if $q_s$ and $q_g$ are not separated by C-obstacle; in other words, they must reside in the same connected component of the C-free space. Conversely, motion planning is infeasible if $q_s$ and $q_g$ are separated by C-obstacle into different connected components. Formally, an infeasibility proof is a closed manifold that entirely resides within C-obstacle and separates the start and the goal~\cite{li2023IJRR}.

The \texttt{SegmentCB} subroutine (Algorithm~\ref{algo:MPI}, line 5) segments $\C{CB}$ into different connected C-free regions separated by C-obstacle. The \texttt{SegmentCheck} function (Algorithm~\ref{algo:MPI}, line 6) verifies whether the start and goal components are separated by C-obstacle. If they are divided into different C-free regions, then motion planning is infeasible. 

To achieve segmentation, we employ an off-the-shelf segmentation function that utilizes a union–find algorithm with a computational complexity of $O(n \log^* n)$, where $n$ is the size of $\C{CB}$. This complexity is effectively near-linear~\cite{sedgewick2004algorithms} and provides substantial computational advantages over traditional search-based methods such as $A^\star$. We experimentally validate this complexity under varying C-space dimensions in Section~\ref{subsec:segmentation}. The segmentation function assigns unique labels to the connected C-free regions. If $q_s$ and $q_g$ receive the same label, it indicates that they belong to the same connected C-free region, and thus, a motion plan exists. However, if they are assigned different labels, it does not immediately imply the absence of a motion plan. This is because, in scenarios involving motion planning for manipulators, it is essential to consider that orientations are defined modulo $2\pi$ (when there are no joint limits). Therefore, when verifying the disconnection between $q_s$ and $q_g$, the \texttt{SegmentCheck} subroutine accounts for the fact that orientation wraps around at $2\pi$ for joints without limits.
\subsection{Analysis of $\C{CB}$ Resolution}
\label{subsec:resolution}
\begin{figure}[t]
\centering
  \subfloat[]{\includegraphics[trim=35 30 38 20,clip,scale=0.34]{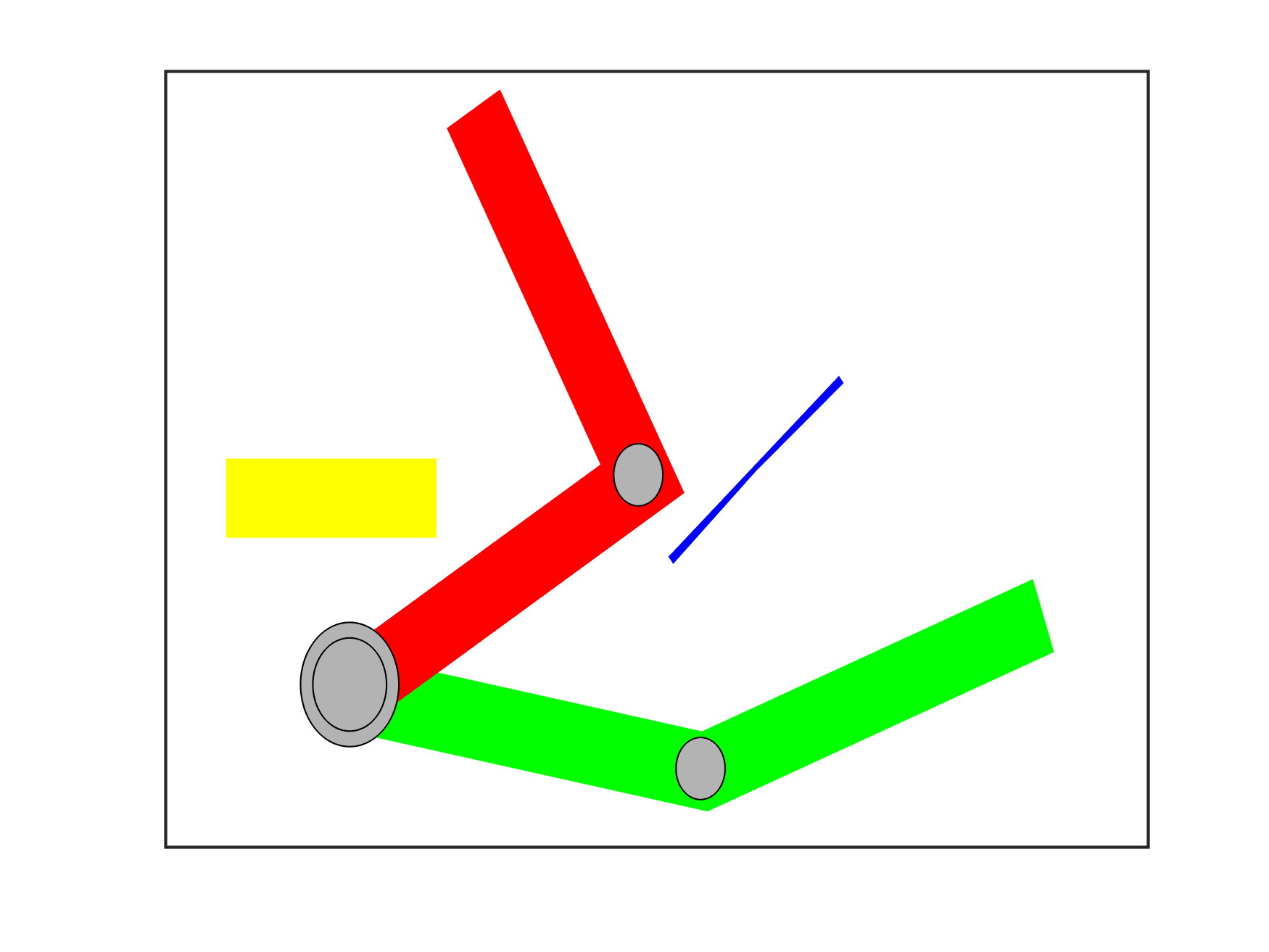}\label{fig:ex}}\hspace{0.02cm}
      \subfloat[]{\includegraphics[trim=93 30 63 10,clip,scale=0.33]{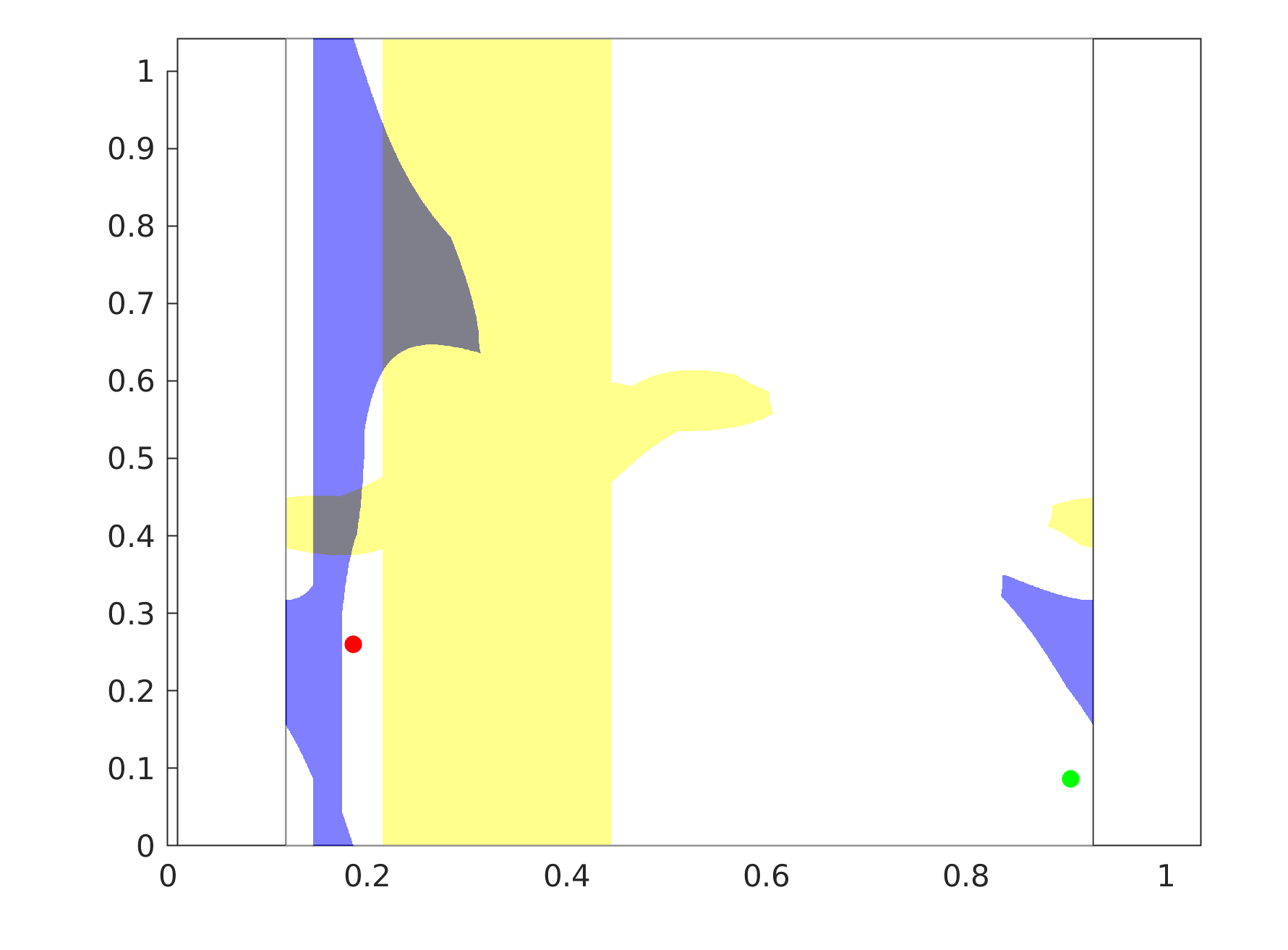}\label{fig:ex1}}\hspace{0.02cm}
                               \subfloat[]{\includegraphics[trim=93 30 63 10,clip,scale=0.33]{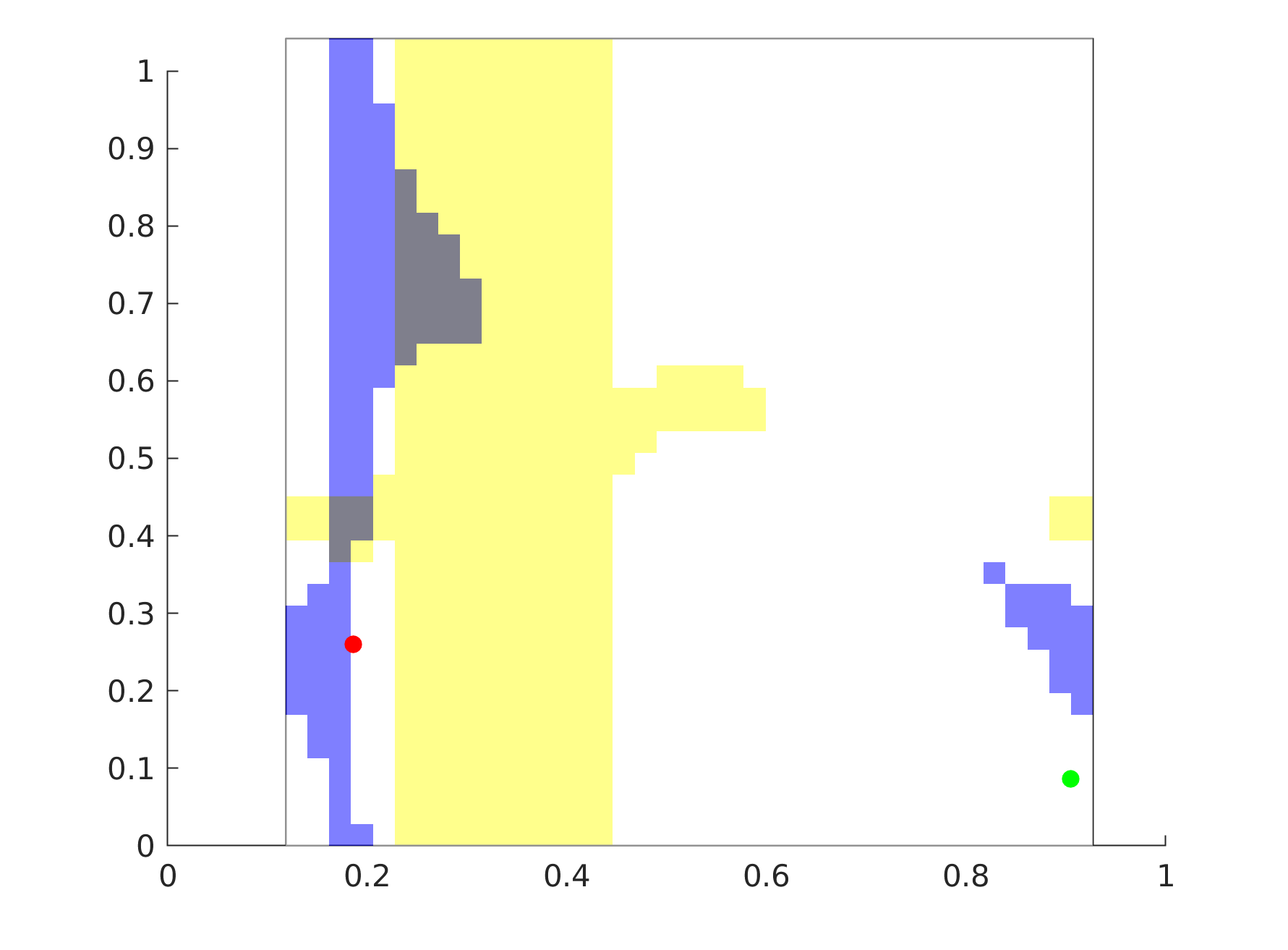}\label{fig:ex2}}\\
                               \subfloat[]{\includegraphics[trim=93 30 63 10,clip,scale=0.33]{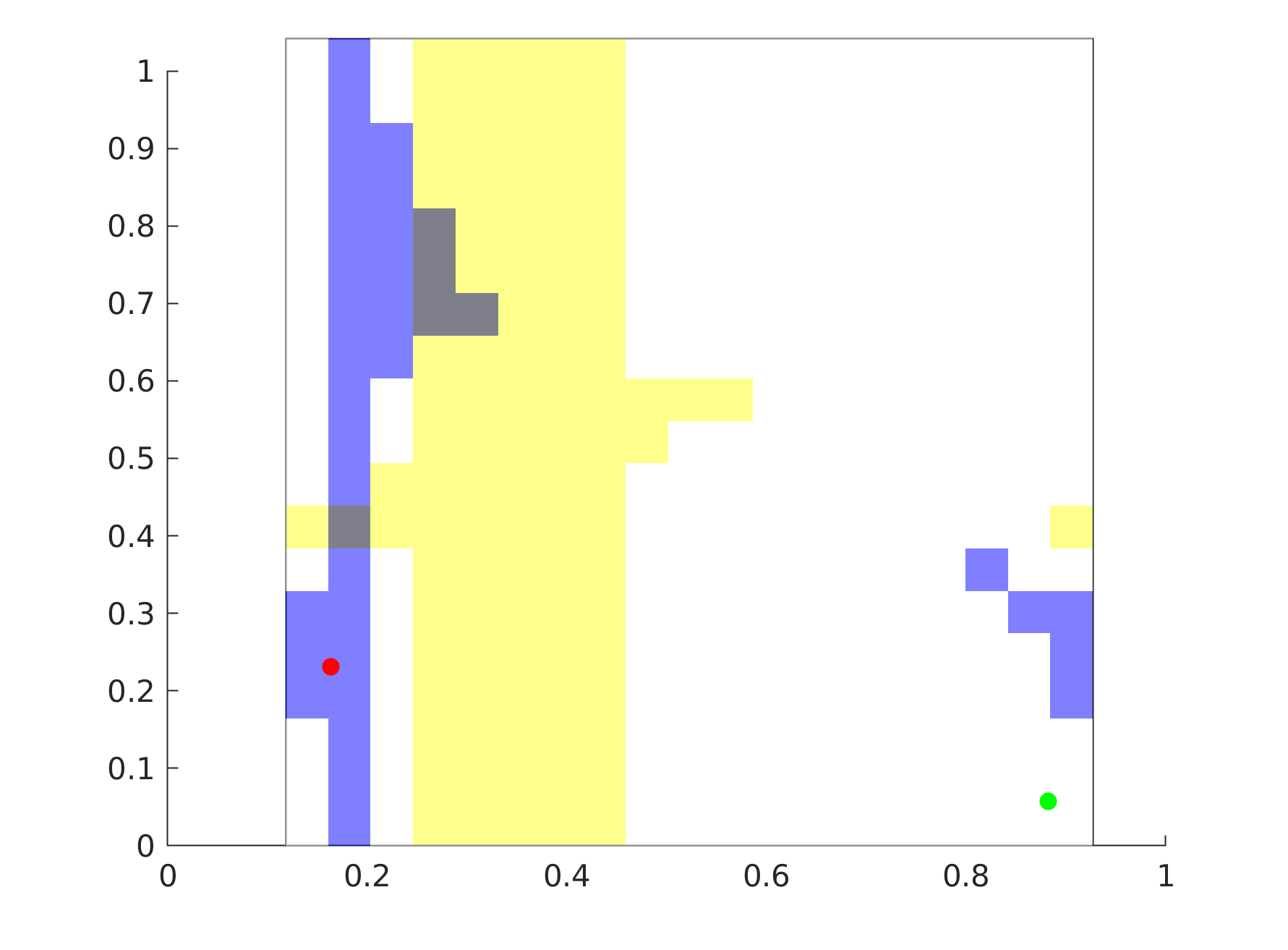}\label{fig:ex3}}\hspace{0.02cm}
                               \subfloat[]{\includegraphics[trim=93 30 63 10,clip,scale=0.33]{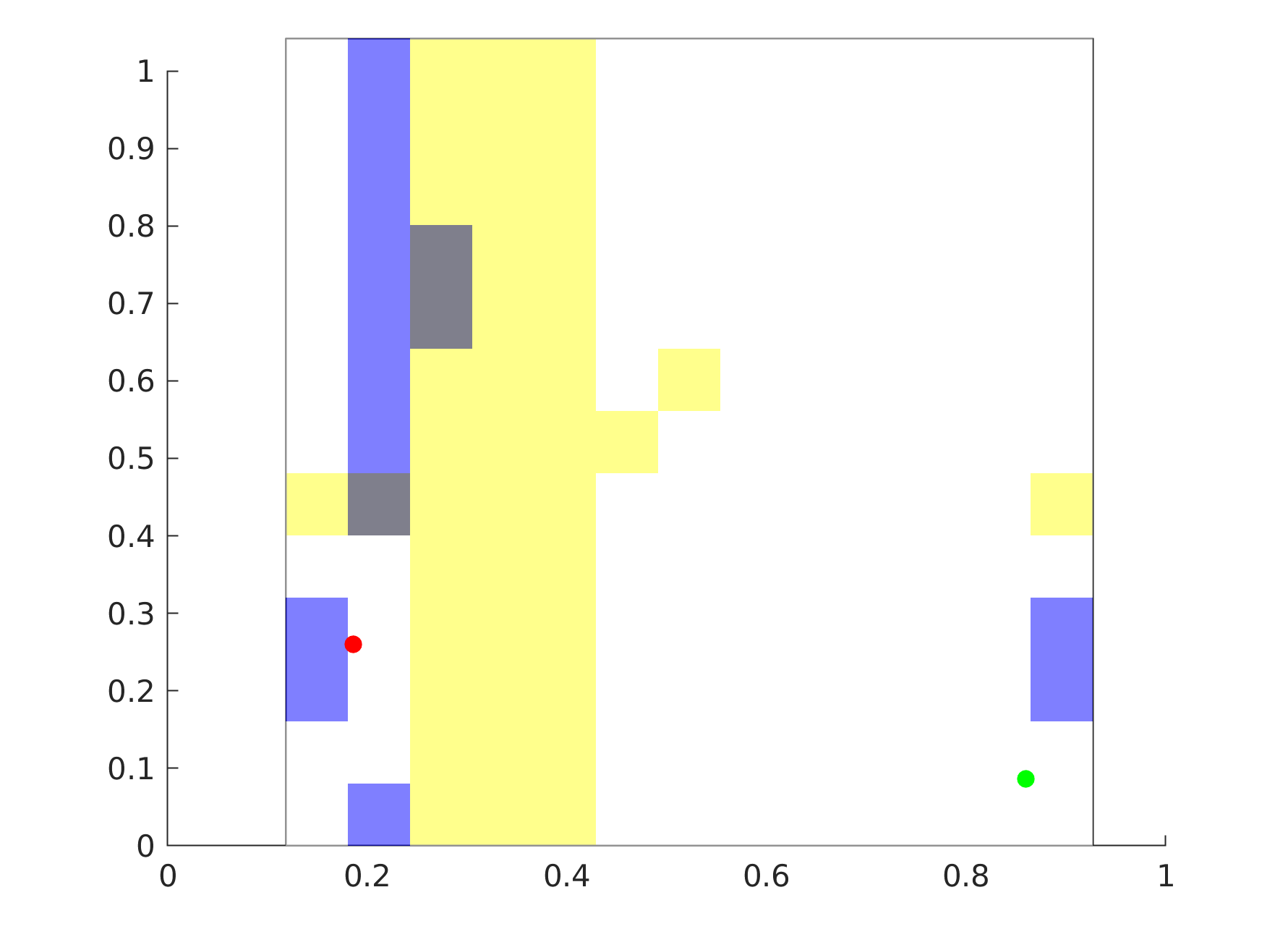}\label{fig:ex4}}
    \caption{(a) 2-link robot in its workspace. (b) The configuration space $\C{C}$ with the start and goal configurations shown as green and red dots, respectively. The different colors represent the correspondence between obstacles in the workspace and obstacles in the C-space. The $\C{C}$ has three connected components. (c) An \textit{equivalent} C-space with $10^\circ$ resolution. (d) C-space with $20^\circ$ resolution. (e) C-space with $30^\circ$ resolution.}
  \label{fig:example}
\end{figure}
The C-obstacle region divides the C-free of the C-space $\mathcal{C}$ into different connected components. For a given $\C{R}$ and $\C{W}$, let us denote the different connected components by $C_1, \ldots, C_m$, whose union results in C-free, denoted as C-free $= \bigcup C_j$. Let the connected components in $\mathcal{CB}$ be denoted by $D_1, \ldots, D_l$. We say that $\mathcal{CB}$ is \textit{equivalent} to $\mathcal{C}$ if
\begin{enumerate}
    \item $l = m$
    \item $D_1 \subset C_1, \ldots, D_l \subset C_m$
    \item $q_s \in \bigcup D_j$
    \item $q_g \in \bigcup D_j$
\end{enumerate}
\begin{theorem}
Let $\C{C}$ be the configuration space corresponding to $\C{R}$ in $\C{W}$. Given $q_s$ and $q_g$ configurations in $\C{C}$ such that motion planning is infeasible, then for any discretized configuration space $\C{CB}$ that is equivalent to $\C{C}$, motion planning remains infeasible for $q_s$ and $q_g$ configurations in $\C{CB}$.
\end{theorem}
\begin{proof}
Let $q_s \in C_i$ and $q_g \in C_j$, so that motion planning is infeasible. From the definition of an equivalent $\mathcal{CB}$, we have $D_k \subset C_k, \forall k\leq l$, $q_s \in \bigcup D_k$, and $q_g \in \bigcup D_k$. Therefore, it follows that $q_s \in D_i$ and $q_g \in D_j$. Thus, no motion plan exists.
\end{proof}
Thus, if motion planning between $q_s$ and $q_g$ remain infeasibile in $\C{C}$, then it remains infeasible in an equivalent $\mathcal{CB}$. 

The concept of equivalent C-space is visualized in Fig.~\ref{fig:example}. A 2-link robot in its workspace, without any joint limits is shown in Fig.~\ref{fig:ex}, where the start and goal states are represented by green and red, respectively. Various obstacles are shown in different colors. The corresponding C-space $\C{C}$ is displayed in Fig.~\ref{fig:ex1}, which illustrates how obstacles in the workspace translate into the C-space. The C-space is discretized at $0.1$ degree for each DOF, resulting in a 3600$\times$3600 bitmap. 

Since there are no joint limits, the C-space topology forms a torus, where the top and bottom edges are connected, as well as the left and right edges. Thus, there are three connected components, and no feasible motion plan exists since $q_s$ and $q_g$ are not in the same connected component. Figure~\ref{fig:ex2} illustrates a $\C{CB}$ of size $36 \times 36$, corresponding to a discretization of $10^\circ$ along each dimension. This bitmap is equivalent to the continuous C-space $\C{C}$. It can be easily verified that there is no collision-free path between $q_s$ and $q_g$ since they are in different connected components. Figure~\ref{fig:ex3} depicts a $\C{CB}$ with a coarser discretization of $20^\circ$ per dimension (resulting in an $18 \times 18$ array). Although the number of connected components remains unchanged, Condition~4 is violated ($q_g$ is incorrectly identified as inside a C-obstacle), and consequently, $\C{CB}$ is not equivalent to $\C{C}$. Finally, Figure~\ref{fig:ex4} shows a $\C{CB}$ with an even lower resolution of $30^\circ$ along each dimension. With only a single connected component, it is clearly not equivalent to $\C{C}$, and wrongly certifies that motion planning is feasible. These examples highlight how a reduction in discretization fidelity can distort the topological structure of $\C{C}$, leading to incorrect connectivity inference in the bitmap representation. This motivates the need for a principled approach to selecting an appropriate resolution that ensures $\C{CB}$ remains topologically consistent with $\C{C}$. 
% One such procedure is described in Section~\ref{subsec:resolution}.

\subsection{Choosing a resolution for $\C{CB}$}
Let us consider a robot with $n$ revolute joints operating in a 3D workspace with known obstacle geometries and locations. 
Let $\C{O}_1, \C{O}_2, \dots$ denote the obstacles, each with a known size $w_{{obs}}^{(k)}$ and pairwise inter-obstacle distances $d_{{gap}}^{(k,\ell)}$ between obstacles $\C{O}_k$ and $\C{O}_\ell$ in the workspace. 
The inter-obstacle distance is defined as the minimal translation required for one obstacle to touch the other.

By abuse of notation, let $\C{R} \subset \mathbb{R}^3$ be the set of all points that constitute the robot’s body in the workspace.
The forward kinematics map $f$ can be defined as
\begin{equation}
f: \C{C} \times \C{R} \rightarrow \mathbb{R}^3,
\quad f(q, x_r) = x
\end{equation}
that is, it maps a configuration $q \in \C{C}$ and a robot body point $x_r \in \C{R}$ to a point $x \in \mathbb{R}^3$ in the workspace. 
Furthermore, let $\C{R}_i \subset \C{R}$ denote the set of points belonging to the $i^\text{th}$ robot link. 
Configuration changes $\Delta q$ can be mapped to corresponding displacements in the workspace $\|\Delta x\|$ as
\begin{equation}
\|\Delta x\| = \|J(q, x_r) \Delta q\|
\end{equation}
where the Jacobian $J(q, x_r) $ is the manipulator Jacobian
\begin{equation}
J(q, x_r) = \frac{\partial f(q, x_r)}{\partial q}
\end{equation}
Therefore, a configuration change $\Delta q$ causes a robot's body point $x_r$ to sweep an arc of length $\|\Delta x\|$. 
To ensure that the discretization map does not allow the robot to sweep over an obstacle (mistaking it for free space), and does not falsely connect distinct free regions separated by obstacles, we must enforce the following requirements.
\begin{enumerate}
\item[($R_1$)] 
For each obstacle $\C{O}_k$ and robot link $i$, the maximum sweep $\|J(q, x_r) \Delta q\|$ (for $x_r \in \C{R}_i$) must be less than the obstacle’s thickness, that is,
\begin{equation}
\|J(q, x_r) \Delta q\| < w_{{obs}}^{(k)}
\end{equation}
\item[($R_2$)] 
For each pairwise obstacles $(\C{O}_k, \C{O}_\ell)$, the maximum sweep must be less than the inter-obstacle gap, that is,
\begin{equation}
\|J(q, x_r) \Delta q\| < d_{{gap}}^{(k,\ell)}
\end{equation}
\end{enumerate}
where $\| \cdot \|$ denotes the 2-norm of a matrix. Since it holds that $\|J(q, x_r) \Delta q\| \leq \|J(q, x_r)\| \cdot \|\Delta q\|$, we obtain that
\begin{equation}
\|\Delta q\| < \frac{\min(w_{\mathrm{obs}}^{(k)}, d_{\mathrm{gap}}^{(k,\ell)})}{\|J(q, x_r)\|}
\end{equation}
By minimizing across all robot points $x_r \in \mathcal{R}$, links $i$, and obstacles $k, \ell$, we can therefore define the resolution
\begin{equation}
\delta^\star = \inf_{q \in \mathcal{C},\; x_r \in \mathcal{R}} \left( \frac{\min(w_{\mathrm{obs}}, d_{\mathrm{gap}})}{\|J(q, x_r)\|} \right)
\label{eq:delta_star}
\end{equation}

Thus, any resolution\footnote{$\delta^\star$ is valid uniformly over all joint angles.} $\delta < \delta^\star$ ensures that $\C{CB}$ faithfully preserves the connectivity structure of the continuous space, preventing the robot from sweeping over obstacles or erroneously jumping across narrow passages. Though~\eqref{eq:delta_star} gives a theoretical framework for computing $\delta^\star$, it requires minimizing a Jacobian-based bound over all robot configurations and all body points, which is a hard optimization problem. Below we give a practical method to compute $\delta^\star$. 
\subsection{Computing $\delta^\star$}
\begin{algorithm}[t]
% \color{revisioncolor}
\caption{Estimation of $\delta^\star$}
\label{alg:estimate_d}
\begin{algorithmic}[1]
\Require{Robot $\C{R}$, obstacles $\C{O}$,samples $S$, $\C{R}_{sub} = \{\C{R}_{1:m} \mid m = 1, \ldots, n \}$}
\Ensure{$\delta^\star$}
\For{each $\C{O}_j \in \C{O}$}
\State{$\C{J}_1 \ldots \C{J}_n \gets \emptyset $}
\For{each $\C{R}_{1:m} \in \C{R}_{sub}$}
        \State{Select $p_l$; sample $\{p_u^{(i)}\}_{i=1}^S$}
        \State{$c_l \leftarrow \mathrm{IK}(p_l)$, $J_l^m \leftarrow J^m(c_l)$}
        \For{$i=1$ to $S$}
            \State{$c_u \leftarrow \mathrm{IK}(p_u^{(i)})$, $J_u^m \leftarrow J^m(c_u)$}
            \State{$\Delta c \leftarrow |c_u-c_l|$}
            \State{$J_{\text{avg}} \leftarrow (J_l^m+J_u^m)/2$}
                    \State{$\C{J}_k \leftarrow \C{J}_k \cup \{\Delta c(k) \|J_{\text{avg}}(:,k)\|\},~\forall k \in \{1, \ldots, m\}$}
            \LineComment \textcolor{mygray}{$\Delta c(k)$ is the $k$-th element of $\Delta c$ and $\|J_{\text{avg}}(:,k)\|$ is the norm of the $k$-th column of $J_{\text{avg}}$.}
                   \EndFor
        \State{$\delta_{\C{O}_j, m} \leftarrow \min \mathcal{J}_k$}
    \EndFor
    \State{$\delta_{\C{O}_j} \leftarrow \max \delta_{\C{O}_j, m}$}
    \EndFor
     \State{$\delta^\star \leftarrow \min_{\C{O}} \delta_{\C{O}_j}$}
\State{\Return{$\delta^\star$}}
\end{algorithmic}
\end{algorithm}

To determine a suitable discretization resolution $\delta$ for the configuration space, we propose a sampling-based procedure that empirically estimates the minimum joint displacement required to traverse each obstacle (or inter-obstacle distance) in the environment. The procedure is summarized in Algorithm~\ref{alg:estimate_d} and described in detail below. 

For each obstacle $\C{O}_i \in \C{O}$, a reference point $p_l$ is selected on a chosen boundary face. In addition, a set of $S$ points $\{p_u^{(i)}\}_{i=1}^S$ are sampled on the obstacle boundary. Note that for primitive shapes it is relatively straightforward to identify the faces corresponding to the minimum dimension of the obstacle which represents the thinnest and most constrained direction of traversal. A fixed reference point $p_l$ is selected on this face, and a set of $S$ are drawn on the opposite face.

\begin{figure}[t]
% \captionsetup{font={color=revisioncolor}}
\centering
  \subfloat[]{\includegraphics[trim=50 05 70 40,clip,scale=0.45]{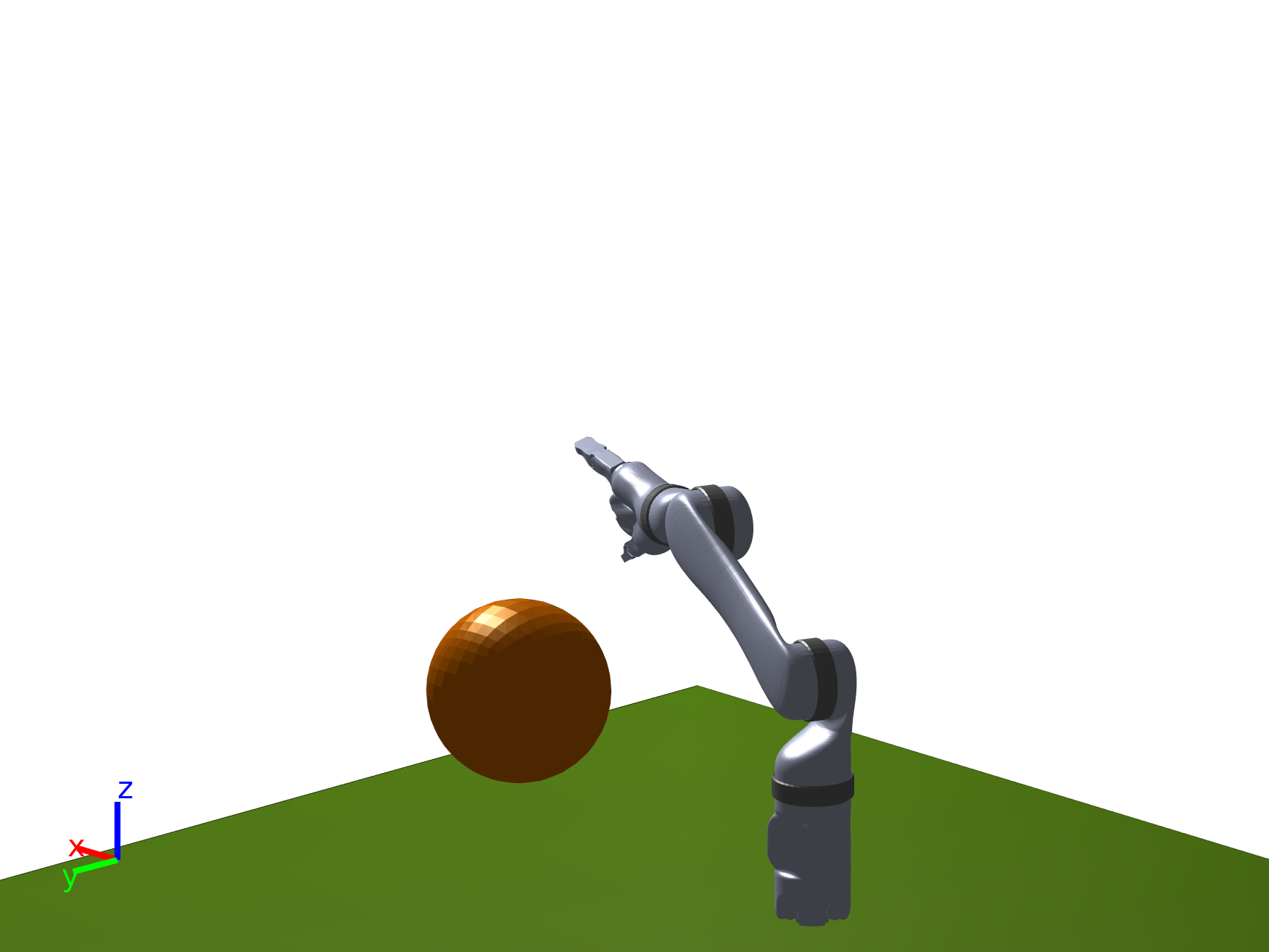}\label{fig:shpere0}} \hspace{0.05cm}
        \subfloat[]{\includegraphics[trim=50 05 35 20,clip,scale=0.45]{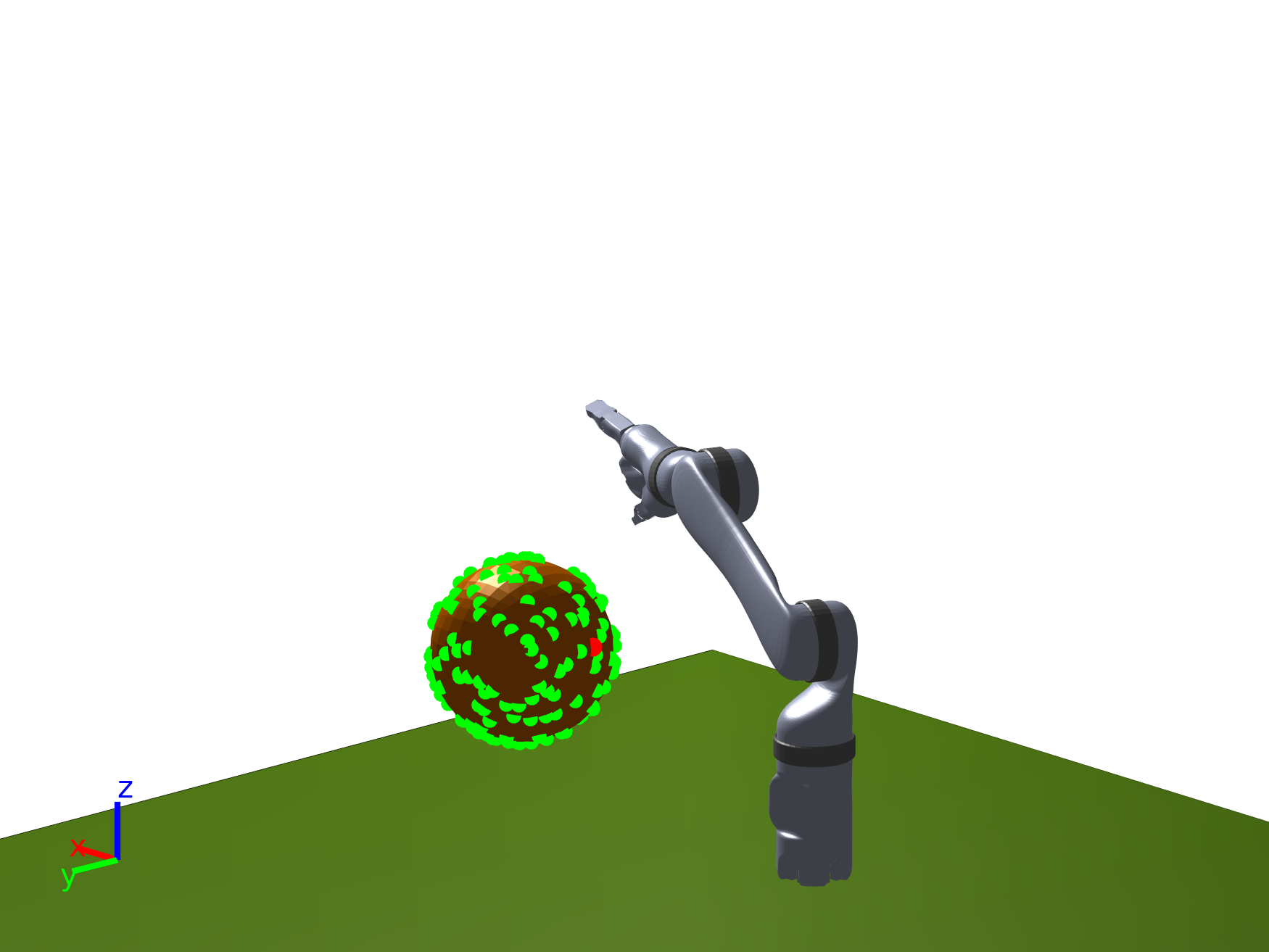}\label{fig:shpere1}}
       \caption{(a) Robot arm and a spherical object. (b) 200 points randomly sampled on the surface of the object. The point $p_l$ is highlighted in red, while the other points are shown in green.}
       \label{fig:delta_compare}
\end{figure}

A progressive strategy is then adopted in which the robot is considered as a sequence of partial chains $\C{R}_{1:m}$, for $m = 1, \dots, n$, where $\C{R}_{1:m}$ denotes the sub-chain comprising the first $m$ links. For each partial chain, inverse kinematics is solved independently for the reference point $p_l$ and each sample $p_u^{(i)}$, yielding configurations $c_l^{(m)}$ and $c_u^{(m)}$ respectively, along with corresponding geometric Jacobians $J_l^{(m)}$ and $J_u^{(m)}$ (lines 5 and 7, Algorithm~\ref{alg:estimate_d}) For each partial chain $\C{R}_{1:m}$ and each sample $p_u^{(i)}$, the joint displacement vector is computed (lines 8, Algorithm~\ref{alg:estimate_d}) which is then scaled by the average Jacobian for the $m$-link chain (line 10), to account for the differing kinematic contributions of each joint to Cartesian motion. The $min$ over $\Delta c(k) \|J_{\text{avg}}(:,k)\|$ identifies $\delta_{\C{O}_j, m}$ (line 11), the most conservative weighted displacement for each joint $m$, and the $max$ over $\delta_{\C{O}_j, m}$ selects the discretization that most meaningfully encodes the obstacle geometry for that chain. Taking the $max$ is motivated by the fact that different links of the robot may interact with the same obstacle differently. Taking the $max$ across $\delta_{\C{O}_j, m}$ identifies the joint whose motion most faithfully encodes the obstacle geometry in Cartesian space, as joints with near-zero weighted displacement are not meaningfully participating in the traversal of that obstacle and should not drive the resolution estimate.\footnote{Note that this may be viewed as a discrete approximation of the norm of the Jacobian in~\eqref{eq:delta_star}.} The required resolution is then obtained as the minimum  $\delta_{\C{O}_j}$ across all the obstacles. 

To compare the resolution estimate of Algorithm~\ref{alg:estimate_d} against a direct numerical evaluation of~\eqref{eq:delta_star}, we consider the scenario shown in Fig.~\ref{fig:delta_compare}, consisting of a robot arm and a spherical obstacle of radius 0.1 m. For Algorithm~\ref{alg:estimate_d}, 200 points are sampled uniformly from the obstacle surface. To numerically evaluate~\eqref{eq:delta_star}, we sample 80,000 random joint configurations and identify those for which at least one link segment geometrically intersects the sphere, determined analytically via the quadratic intersection of the parametric link segment with the sphere. For each intersecting link, the entry and exit points with the obstacle surface are computed, and 50 points are sampled uniformly along the chord between them. This is done since the Jacobian $J(q, x_r)$ varies along the link interior, and the largest $\|J(q,x_r)\|$, may occur at an interior point rather than at the link endpoint frames.

Algorithm~\ref{alg:estimate_d} yields a value of $\delta^\star = 7.54^\circ$ and completes in 15 seconds on average, whereas the numerical implementation of~\eqref{eq:delta_star} yields $\delta^\star = 6.97^\circ$ and requires approximately 170 seconds.

\section{Experimental Analysis and Discussion}
\label{sec:experiments}
\begin{figure}[t!]
% \captionsetup{font={color=revisioncolor}}
\centering
  \subfloat[]{\includegraphics[trim=58 55 70 40,clip,scale=0.45]{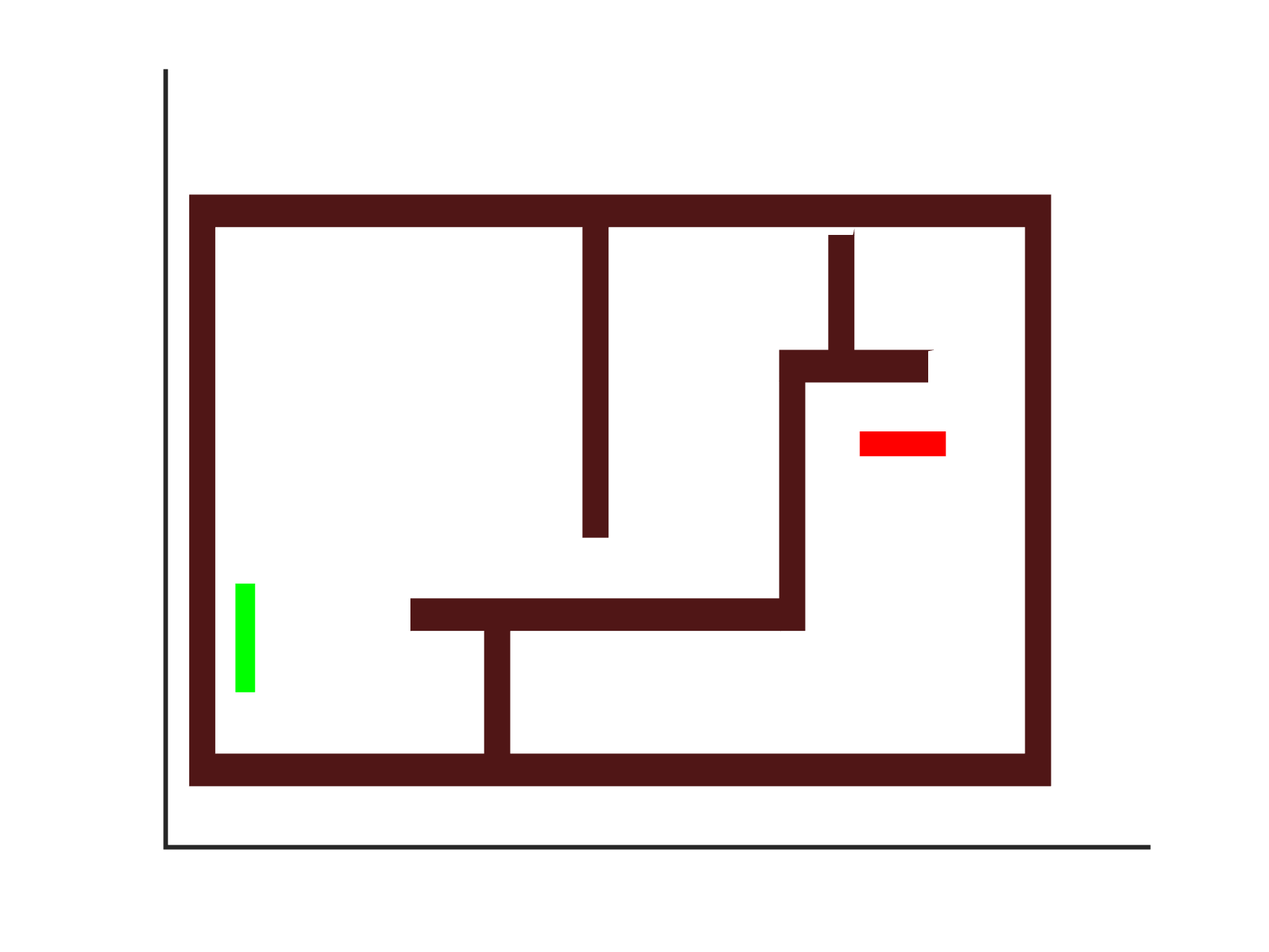}\label{fig:3dof}}\\
        \subfloat[]{\includegraphics[trim=50 35 35 20,clip,scale=0.455]{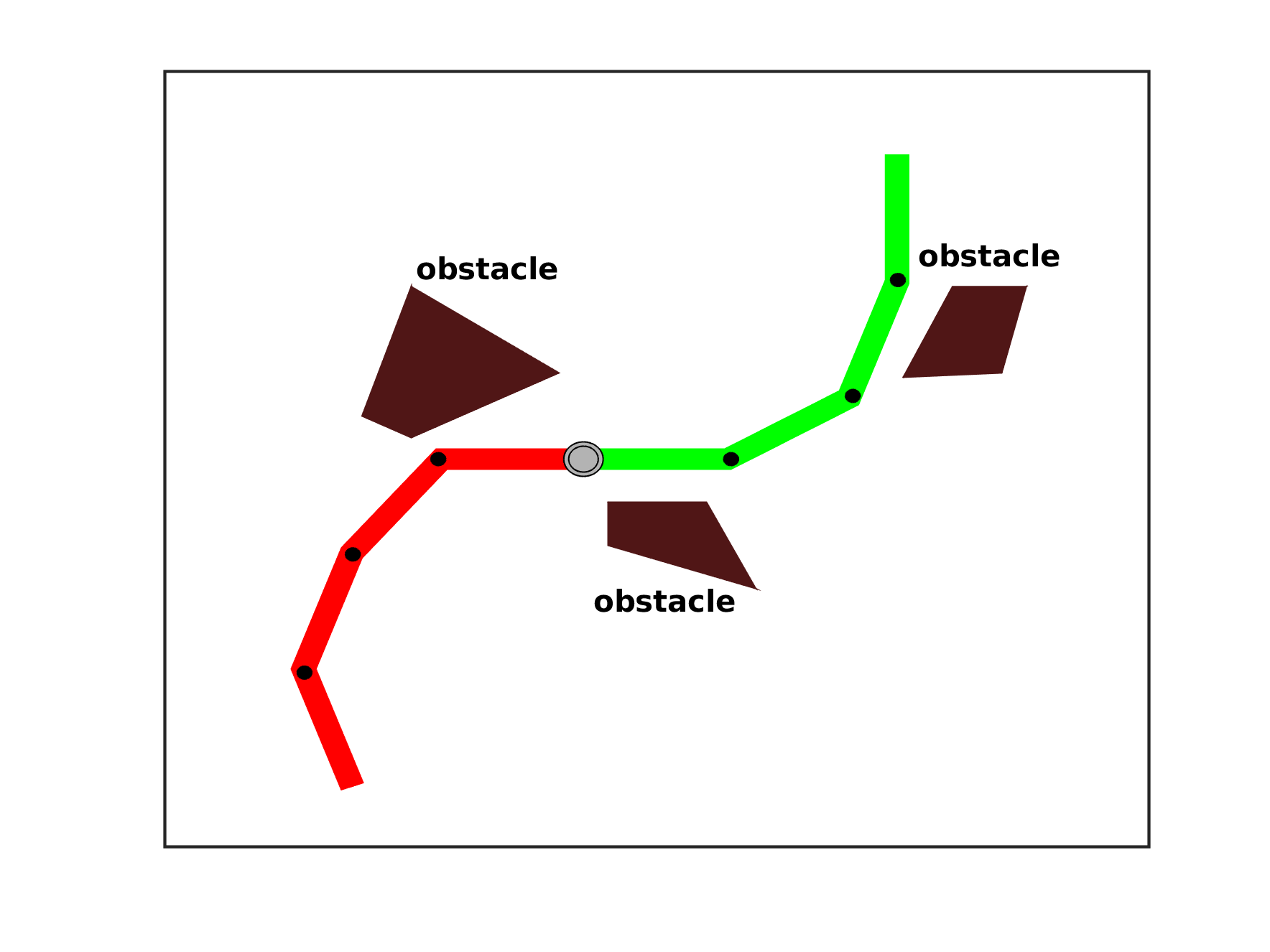}\label{fig:4dof}}
                               \subfloat[]{\includegraphics[trim=50 32 35 20,clip,scale=0.4495]{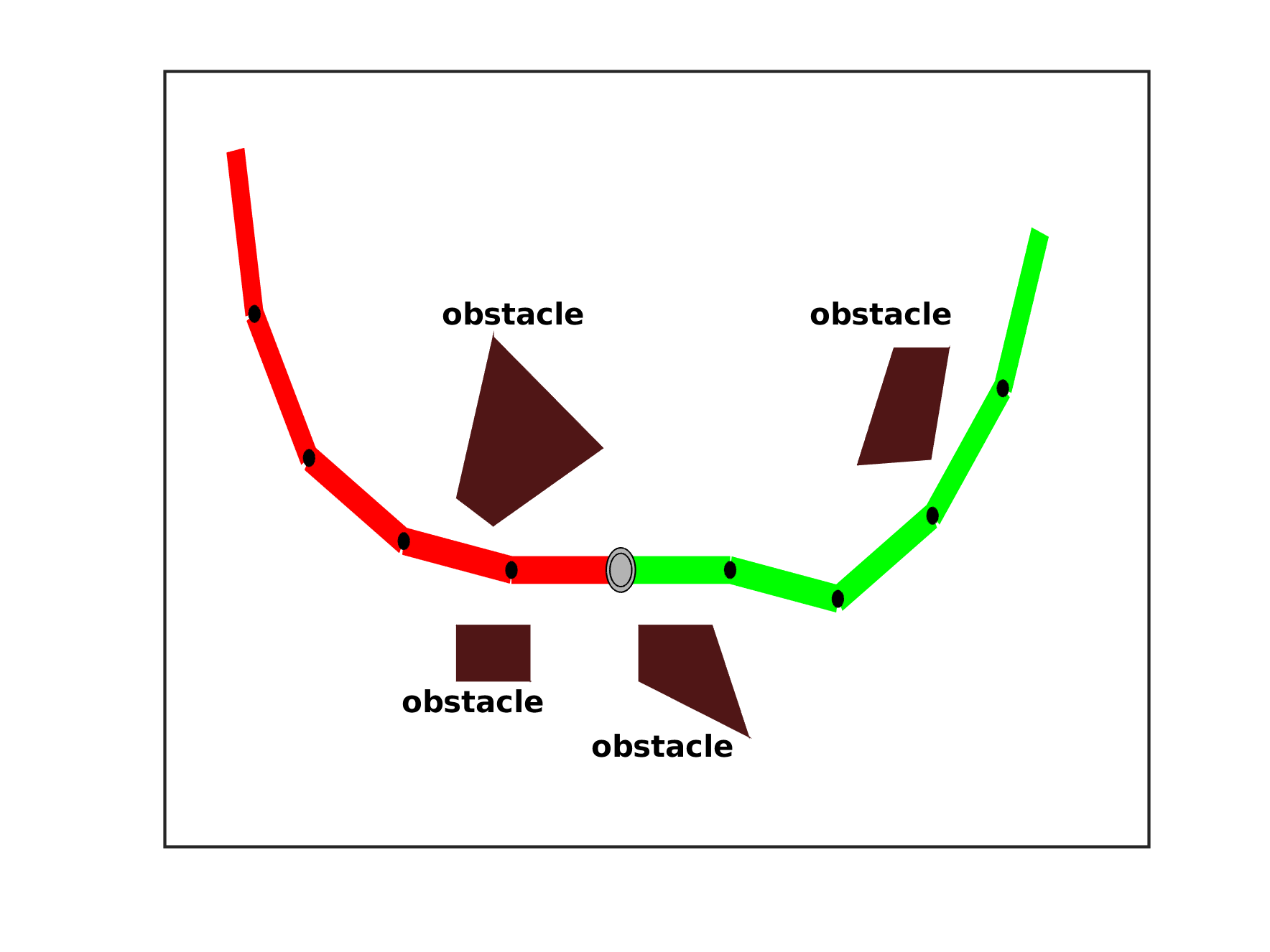}\label{fig:5dof}}
   \caption{Different types of robots used in the 2D experiments. The start configuration of each robot is shown in \textcolor{green}{green}, and the goal configuration is shown in \textcolor{red}{red}. The obstacle regions are depicted in \textcolor{mybrown}{brown}. (a) $S_1$: 3-DOF rectangular robot scene. (b) $S_2$: 4-DOF articulated robot scene. (c) $S_5$: 5-DOF articulated robot scene.}

  \label{fig:experiments}
\end{figure}
% \subfloat[]{\includegraphics[trim=120 150 100 20,clip,scale=0.6]{figures/kinova1}\label{fig:3D1}}
%      \subfloat[]{\includegraphics[trim=140 150 60 20,clip,scale=0.6]{figures/kinova2}\label{fig:3D2}}
\begin{figure}[]
\centering
  \subfloat[]{\includegraphics[trim=120 150 90 20,clip,scale=0.75]{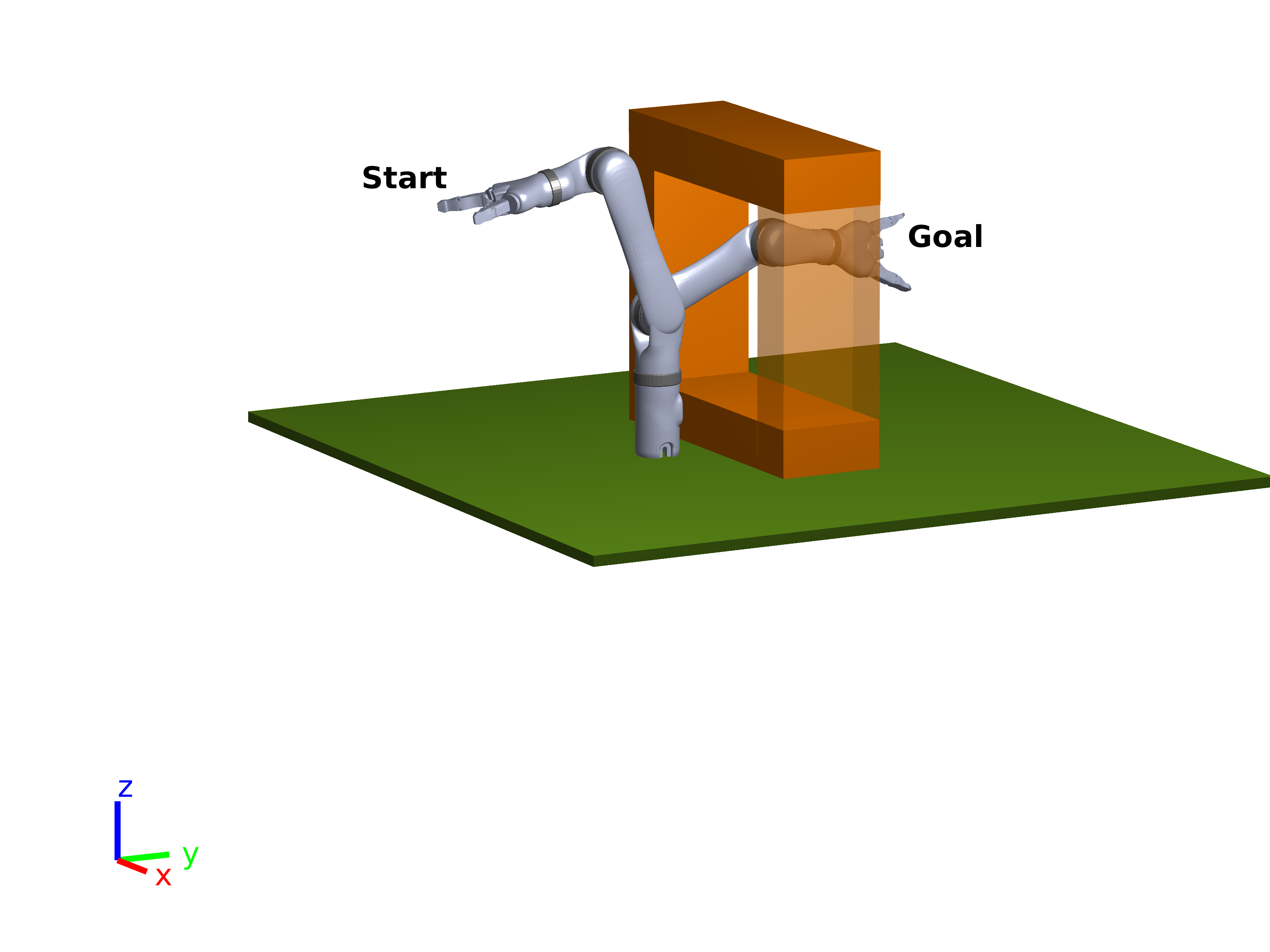}\label{fig:3D1}}\hfill
      \subfloat[]{\includegraphics[trim=140 150 70 20,clip,scale=0.75]{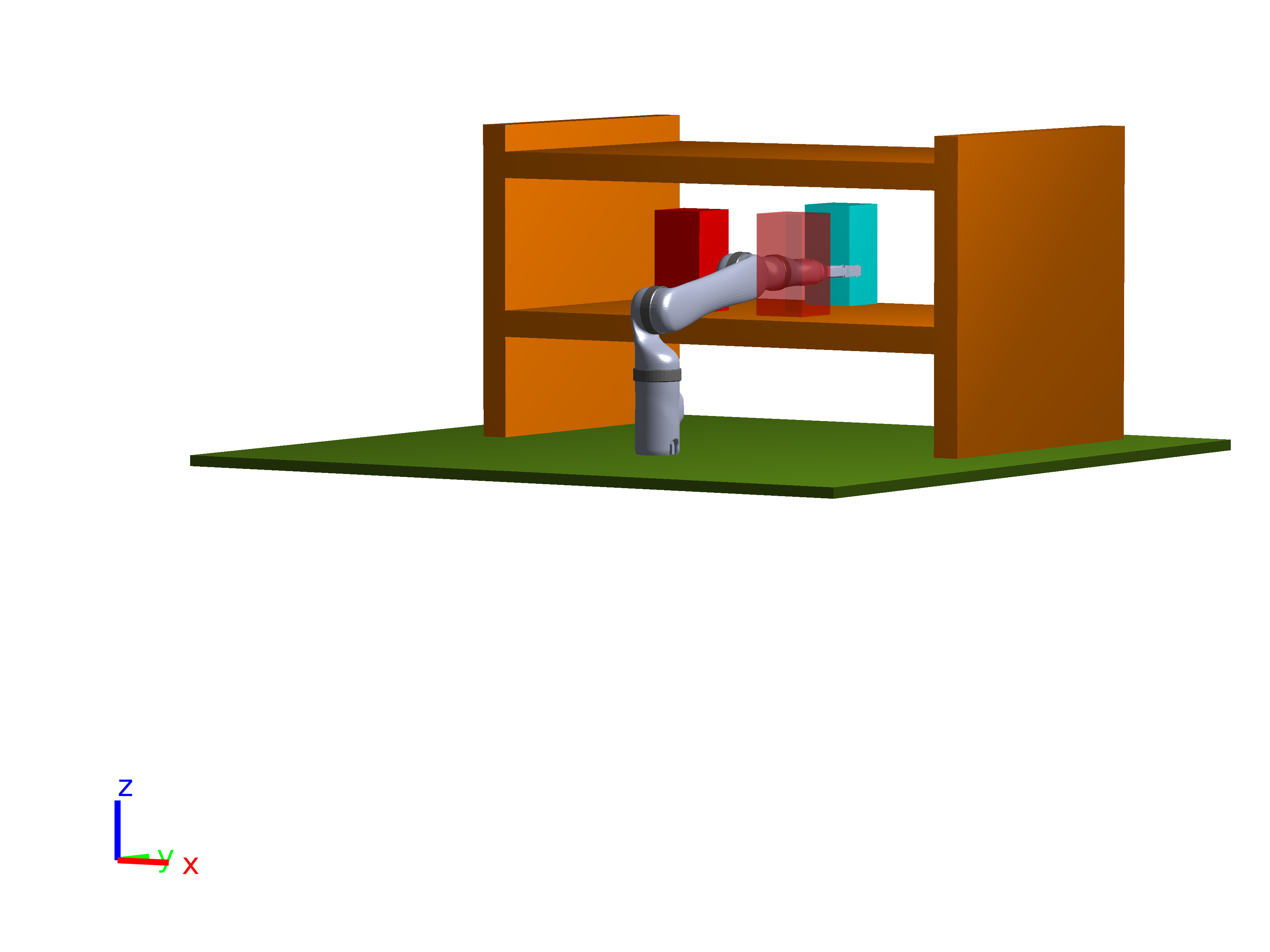}\label{fig:3D2}}\\
      \subfloat[]{\includegraphics[trim=100 150 70 0,clip,scale=0.7]{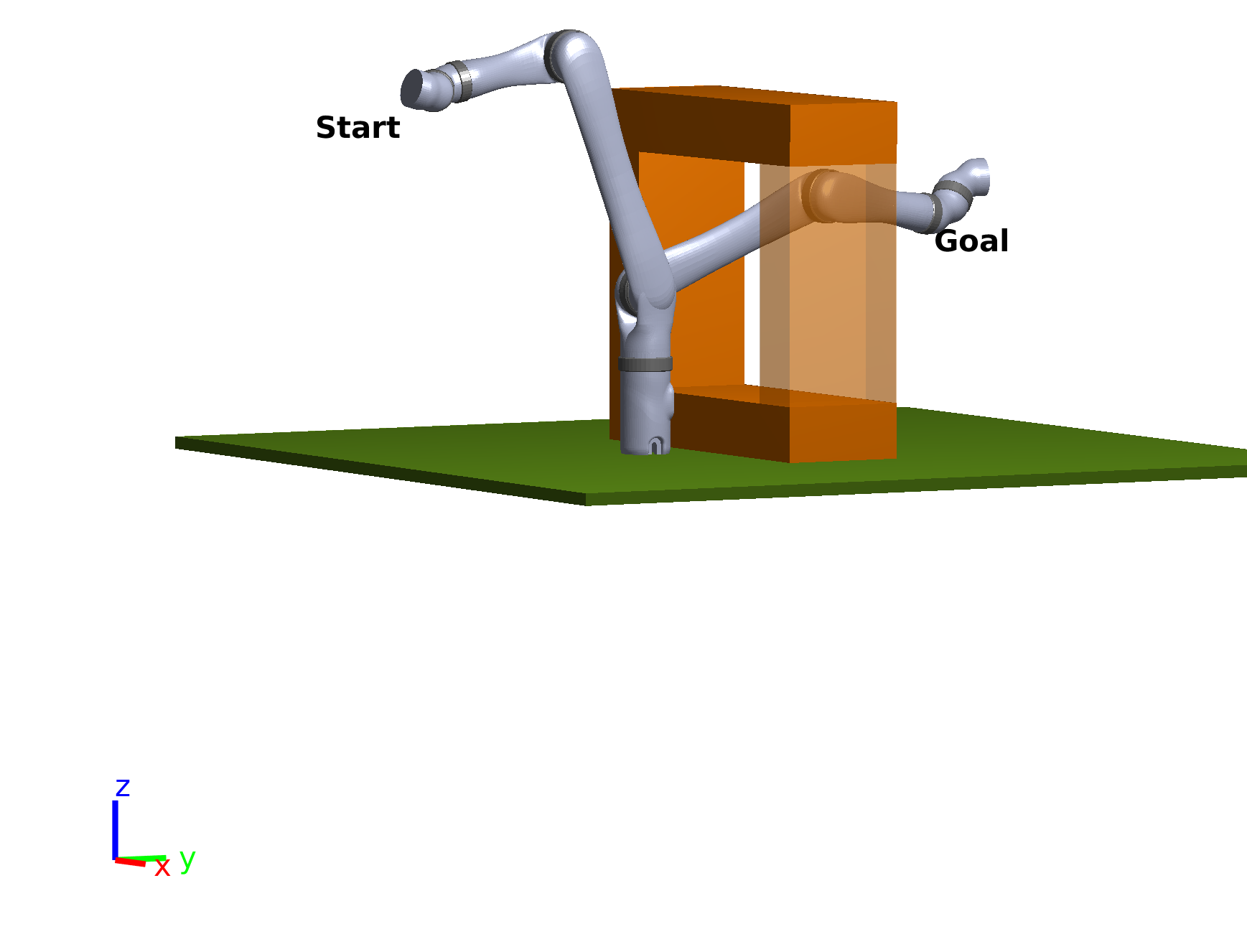}\label{fig:3D_5dof}}
                    \caption{Experiment scenes involving the 4-DOF Kinova arm: (a) $S_3$: The Kinova arm attempting to reach inside the frame. (b) $S_4$: The robot arm trying to grasp the cyan block from a position outside the shelf. The red blocks segment the start and goal configurations into different connected components, preventing any path. (c) $S_6$: A 5-DOF robot attempting to reach inside the frame.}
  \label{fig:experiments2}
\end{figure}
We conduct experiments on six different motion planning problems in both 2D and 3D workspaces. The 3D workspace scenarios, inspired by \cite{li2023IJRR}, are performed using the 4-DOF Kinova MICO arm. The scenarios are as follows: 
\begin{itemize}
    \item[($S_1$)] A scenario involving a 2D robot capable of translation and rotation (see Fig.~\ref{fig:3dof}),
    \item[($S_2$)] A scenario with a 4-DOF articulated robot, as depicted in Fig.~\ref{fig:4dof},
    \item[($S_3$)] A scenario where the Kinova arm is trying to reach inside a frame (see Fig.~\ref{fig:3D1}),
    \item[($S_4$)] A scenario with the Kinova arm attempting to grasp the cyan block from a position outside the shelf, as shown in Fig.~\ref{fig:3D2},
    \item[($S_5$)] A scenario involving a 5-DOF articulated robot aiming to reach a goal configuration (Fig.~\ref{fig:5dof}),
    \item[($S_6$)] A 5-DOF robot (Fig.~\ref{fig:3D_5dof}) trying to reach inside a frame from a position outside it.
\end{itemize}
\begin{table}[]
\begin{center}
\scalebox{0.7}{
\begin{tabular}{c c c c c c c c} 
\hline
\rule{0pt}{1.005\normalbaselineskip}
Scenarios & DOF & C-space size & $ns$ & $d$ & Iterations & Segmentation time (s) & Total time (s) \\
\hline
\hline
\rule{0pt}{1.005\normalbaselineskip}
\multirow{3}{*}{$S_1$ (Fig.~\ref{fig:3dof})} & \multirow{3}{*}{3} & 34$\times$20$\times$36 & \multirow{3}{*}{100} & \multirow{3}{*}{5} & 11.76$\pm$4.18 & 0.00$\pm$0.00 & 2.71$\pm$0.94\\
 &  & 45$\times$36$\times$48 & & & 17.10$\pm$6.59 & 0.00$\pm$0.00 & 4.19$\pm$1.45\\
 & & 67$\times$39$\times$72 & & & 28.33$\pm$11.13 & 0.01$\pm$0.00 & 5.19$\pm$1.93 \\
 \hline
 \rule{0pt}{1.005\normalbaselineskip}
 \multirow{3}{*}{$S_2$ (Fig.~\ref{fig:4dof})} & \multirow{3}{*}{4} &36$\times$36$\times$36$\times$36 & \multirow{3}{*}{100} & \multirow{3}{*}{5} & 1.00$\pm$0.00 & 0.09$\pm$0.00 & 0.62$\pm$0.06 \\
 &  & 48$\times$48$\times$48$\times$48 & & & 1.00$\pm$0.00 & 0.31$\pm$0.02 & 0.89$\pm$0.07\\
 &  & 72$\times$72$\times$72$\times$72 & & & 1.00$\pm$0.00 & 1.74$\pm$0.10 & 2.56$\pm$0.12\\
 \hline
  \rule{0pt}{1.005\normalbaselineskip}
  \multirow{3}{*}{$S_3$ (Fig.~\ref{fig:3D1})} & \multirow{3}{*}{4} & 36$\times$36$\times$36$\times$36 & \multirow{3}{*}{100} & \multirow{3}{*}{5} & 1.47$\pm$0.51 & 0.06$\pm$0.00 & 24.52$\pm$9.18 \\
   &  & 48$\times$48$\times$48$\times$48 & & & 2.43$\pm$0.62 & 0.20$\pm$0.01 & 42.93$\pm$12.42\\
 &  & 72$\times$72$\times$72$\times$72 & & & 3.57$\pm$0.68 & 1.11$\pm$0.05 & 107.27$\pm$19.99\\
 \hline
  \rule{0pt}{1.005\normalbaselineskip}
  \multirow{3}{*}{$S_4$ (Fig.~\ref{fig:3D2})} & \multirow{3}{*}{4} & 36$\times$36$\times$36$\times$36 & \multirow{3}{*}{100} & \multirow{3}{*}{5} & 1.20$\pm$0.48 & 0.07$\pm$0.00 & 28.10$\pm$12.87 \\
   &  & 48$\times$48$\times$48$\times$48 & & & 2.20$\pm$0.85 & 0.21$\pm$0.01 & 56.31$\pm$25.24\\
 &  & 72$\times$72$\times$72$\times$72 & & & 3.43$\pm$1.16 & 1.13$\pm$0.02 & 136.01$\pm$49.23\\
  \hline
   \rule{0pt}{1.005\normalbaselineskip}
 \multirow{3}{*}{$S_5$ (Fig.~\ref{fig:5dof})} & \multirow{3}{*}{5} & 36$\times$36$\times$36$\times$36$\times$36 & \multirow{3}{*}{100} & \multirow{3}{*}{5} & 1.00$\pm$0.00 & 4.43$\pm$0.23 & 5.67$\pm$0.22 \\
  &  & 48$\times$48$\times$48$\times$48$\times$48 & & & 1.00$\pm$0.00 & 20.94$\pm$1.10 & 23.55$\pm$0.97\\
 &  & 72$\times$72$\times$72$\times$72$\times$72 & & & 1.00$\pm$0.00 & 224.46$\pm$5.13 & 255.82$\pm$4.51\\
  \hline
  \rule{0pt}{1.005\normalbaselineskip}
 \multirow{3}{*}{$S_6$ (Fig.~\ref{fig:3D_5dof})} & \multirow{3}{*}{5} & 36$\times$36$\times$36$\times$36$\times$36 & 100 & 5 & 1.26$\pm$0.45 & 4.04$\pm$0.22 & 18.50$\pm$7.38 \\
  &  & 48$\times$48$\times$48$\times$48$\times$48 & 100 & 5 & 1.83$\pm$0.46 & 15.82$\pm$0.56 & 60.92$\pm$15.03\\
 &  & 72$\times$72$\times$72$\times$72$\times$72 & 300 & 5 & 1.00$\pm$0.00 & 215.48$\pm$31.35 & 364.22$\pm$24.57\\
  \hline
 \end{tabular}}
\end{center}
\caption{Motion planning infeasibilty results. \textit{C-space size} denotes the total number of cells in $\C{CB}$. For scenarios $S_2$–$S_5$, this corresponds to resolutions of $10^\circ$, $7.5^\circ$, and $5^\circ$, respectively. For scenario $S_1$, the resolutions correspond to $(1,\text{unit}, 1,\text{unit}, 10^\circ)$, $(0.75,\text{unit}, 0.75,\text{unit}, 7.5^\circ)$, and $(0.5,\text{unit}, 0.5,\text{unit}, 5^\circ)$ for $(x, y, \theta)$, respectively.} \textit{Iterations} denotes the average number of iterations to determine motion planning infeasibility. \textit{Segmentation time} denotes the total execution time of the \texttt{SegmentCB} subroutine, aggregated over all iterations. \textit{Total time} is the overall time taken to report plan infeasibilty.
\label{tab:statistics}
\end{table}

For all scenarios, we empirically set $ns = 100$ and $d = 5$. We recall that $ns$ denotes the minimum number of samples generated by the \texttt{SampleCobstacle} subroutine in a single iteration of Algorithm~\ref{algo:MPI}, while $d$ represents the number of nearby configurations examined for collision.  A detailed justification for these parameter choices will be provided later. Collision detection for the Kinova MICO arm is conducted using the \textit{checkCollision}\footnote{Web: \url{https://it.mathworks.com/help/robotics/ref/rigidbodytree.checkcollision.html}} feature of the MATLAB Robotics and Autonomous Systems toolbox. However, for the 2D scenarios, both the robot and obstacle polygons are subdivided into triangles, and a triangle intersection algorithm~\cite{mccoid2022TMS} is utilized for collision detection. The segmentation of $\C{CB}$ in the \texttt{SegmentCB} subroutine is accomplished by utilizing the MATLAB function \textit{bwlabeln}\footnote{Web: \url{https://it.mathworks.com/help/images/ref/bwlabeln.html}}. This function returns a labeled matrix, assigning labels to different connected components. The performance is evaluated on a Lenovo ThinkPad laptop equipped with an Intel{\small\textregistered} Core i7-10510U CPU$@$1.80GHz$\times$8 with 16GB RAM under Ubuntu 18.04 LTS.

The experimental scenario corresponds to the non-existence of a path between the start and the goal configurations. To establish the ground truth for plan infeasibility, for the Kinova arm scenarios, we run RRT-Connect~\cite{kuffner2000ICRA} continuously for more than 30 minutes, and for the other three scenarios, we run PRM~\cite{kavraki1996IEEE} for more than 30 minutes. For each experiment, we conduct 30 trials and record the following metrics:
\begin{itemize}  
\item \textit{Iterations}: The average number of iterations required to establish motion planning infeasibility. Each iteration of Algorithm~\ref{algo:MPI} involves sampling at least $ns$ configurations within the C-obstacle. Depending on the robot's geometry and the spatial distribution of obstacles in the workspace, a single iteration may not yield sufficient samples within the C-obstacle to successfully segment $q_s$ and $q_g$ into distinct connected components. Consequently, multiple iterations may be necessary to certify infeasibility.  

\item \textit{Segmentation time}: The cumulative execution time of the \texttt{SegmentCB} subroutine, summed across all iterations.  

\item \textit{Total time}: The overall computational time required by Algorithm~\ref{algo:MPI} to ascertain plan infeasibility. This includes the segmentation time.  
\end{itemize}  
Table~\ref{tab:statistics} reports the above metrics. To ensure that the chosen $\C{CB}$ is equivalent to the respective continuous $C$-spaces, we run Algorithm~\ref{alg:estimate_d} to compute $\delta^\star$. The resulting values of $\delta^\star$ for $S_2$ to $S_6$ are $41.79^\circ$, $15.44^\circ$, $15.25^\circ$, $23.51^\circ$, and $16.72^\circ$, respectively. In our experiments, however, we use finer resolutions of $10^\circ$, $7.5^\circ$, and $5^\circ$. Performance is evaluated across these resolutions to assess the associated computational cost.

\subsection{3-DOF scenario}
Infeasibility is detected within a few seconds for all the three resolutions of $\C{CB}$. It is worth noting that the segmentation of the $\C{CB}$ into connected components takes only a fraction of a second for all three resolutions reported. We can conclude that the approach is applicable to 3-DOF robots.

\subsection{4-DOF experiments} 
The scenarios depicted in Fig.~\ref{fig:4dof} ($S_2$) and Fig.~\ref{fig:experiments2} ($S_3$, $S_4$) correspond to a four-dimensional C-space. For $S_2$, as shown in Table~\ref{tab:statistics}, the total runtime remains below 3 seconds across all three resolutions of $\C{CB}$, with infeasibility being established in the first iteration. Due to the sparse distribution of obstacles, each sampled configuration $q$ in C-obstacle, when processed using the \texttt{SpeedUp} subroutine and analyzed along with its $d$ neighboring configurations, quickly identifies C-obstacle configurations contributing to infeasibility. 

Although the scenarios $S_3$ and $S_4$ also correspond to a 4-DOF robot, their total computational time is significantly higher compared to $S_2$. This increase is primarily attributed to the differences in the \texttt{CollisionCheck} subroutine: in the 2D scenario $S_2$, triangle intersection tests employed are computationally efficient, whereas for $S_3$ and $S_4$, the \textit{checkCollision} function introduces a substantially higher computational overhead. 

Comparing $S_3$ and $S_4$, we observe that $S_4$ incurs a slightly higher computational cost. In $S_3$, the frame contributes to the infeasibility, whereas in $S_4$, the two red blocks obstruct the robotic arm from reaching the target position. This implies that if the red blocks are removed from the workspace, a feasible path exists. The current random sampling strategy generates redundant samples from the C-obstacle region contributed by the shelf, which, while contributing to the analysis, are not necessarily required for proving infeasibility. This results in unnecessary computations and increases the overall runtime. Overall, the experiments demonstrate that the proposed approach is well-suited for motion planning in 4-DOF robotic systems.  

\subsection{5-DOF scenario} The proposed approach demonstrates effective scalability to a 5-dimensional C-space, as evidenced by the last two rows of Table~\ref{tab:statistics}. Compared to $S_5$, scenario $S_6$ exhibits an approximately threefold increase in computation time for the $36$ and $48$ resolution cases. Similar to the 4-DOF scenarios, this increase is primarily attributed to the higher cost of the \texttt{CollisionCheck} subroutine in 3D environments relative to 2D.

Furthermore, as the bitmap resolution increases to $72$, the segmentation stage dominates the overall computation time. Consequently, minimizing the number of iterations becomes important, as multiple iterations would significantly increase the total runtime. With $ns = 100$ for $S_6$, the method requires approximately three iterations on average. Increasing the sample count to $ns = 300$ reduces this to a single iteration, but at the expense of additional sampling cost, thereby contributing to the overall computation time.
\section{Discussion}
\label{sec:discussion}
\subsection{Advantages of Segmentation over Search-Based Methods}
\label{subsec:segmentation}
\begin{figure}[]
% \captionsetup{font={color=revisioncolor}}
\centering
  \subfloat[]{\includegraphics[width=0.48\textwidth]{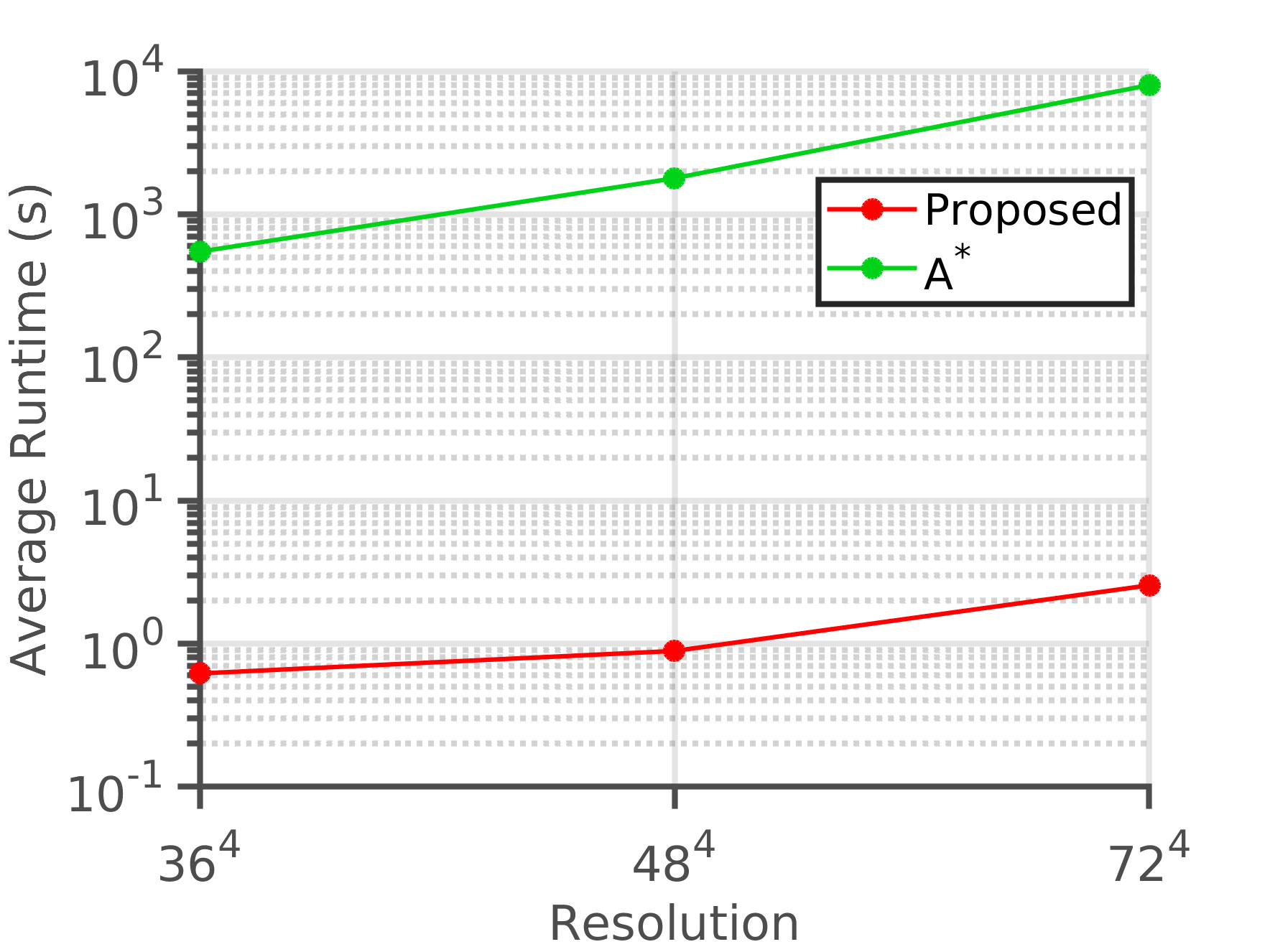}\label{fig:astar1}}\hspace{0.1cm}
      \subfloat[]{\includegraphics[width=0.48\textwidth]{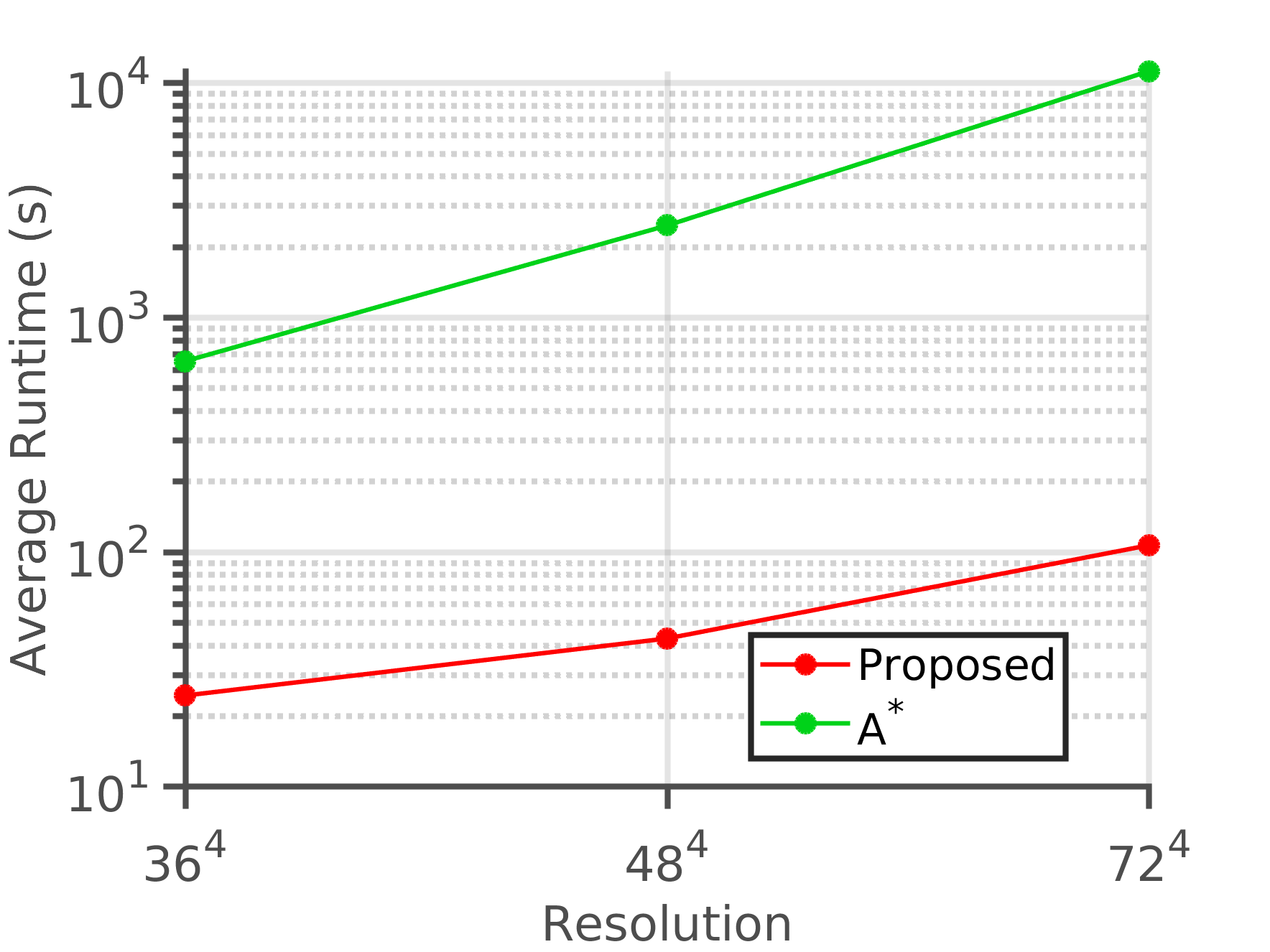}\label{fig:astar2}}\hspace{0.1cm}
                    \caption{Computation times for segmentation-based (proposed) vs. $A^\star$-based infeasibility checks for scenarios (a) $S_2$ and (b) $S_3$.}
                     \label{fig:astar}
\end{figure}
One may argue that, given a discretized C-space, traditional graph-based search algorithms such as $A^\star$ could be employed directly to certify infeasibility, instead of segmenting the C-space into connected components. However, search-based methods must exhaustively explore all possible nodes in the absence of a feasible path, leading to significantly higher computational overhead compared to our segmentation approach.

To provide a baseline comparison, we implemented an $A^\star$-based infeasibility check by replacing the \texttt{SegmentCB} subroutine in Line 5 of Algorithm~\ref{algo:MPI} with a standard $A^\star$ search. If the search terminates without finding a path from $q_s$ to $q_g$, the problem is deemed infeasible. The comparison was performed for scenarios $S_2$ and $S_3$, and the corresponding computation times for the proposed segmentation-based method and the $A^\star$ baseline are shown in Fig.~\ref{fig:astar}. 

For scenario $S_2$, the proposed method required between 0.6-2.6 seconds, whereas the $A^\star$ approach took 550-8000 seconds. Similarly, for scenario $S_3$, the proposed method completed within 25-107 seconds, while the $A^\star$ search required 650-11,000 seconds. These results demonstrate that the proposed segmentation-based approach is approximately \textit{two to four} orders of magnitude faster than the $A^\star$-based baseline across the tested environments.

\begin{figure}[]
% \captionsetup{font={color=revisioncolor}}
\centering
 \includegraphics[width=0.65\textwidth]{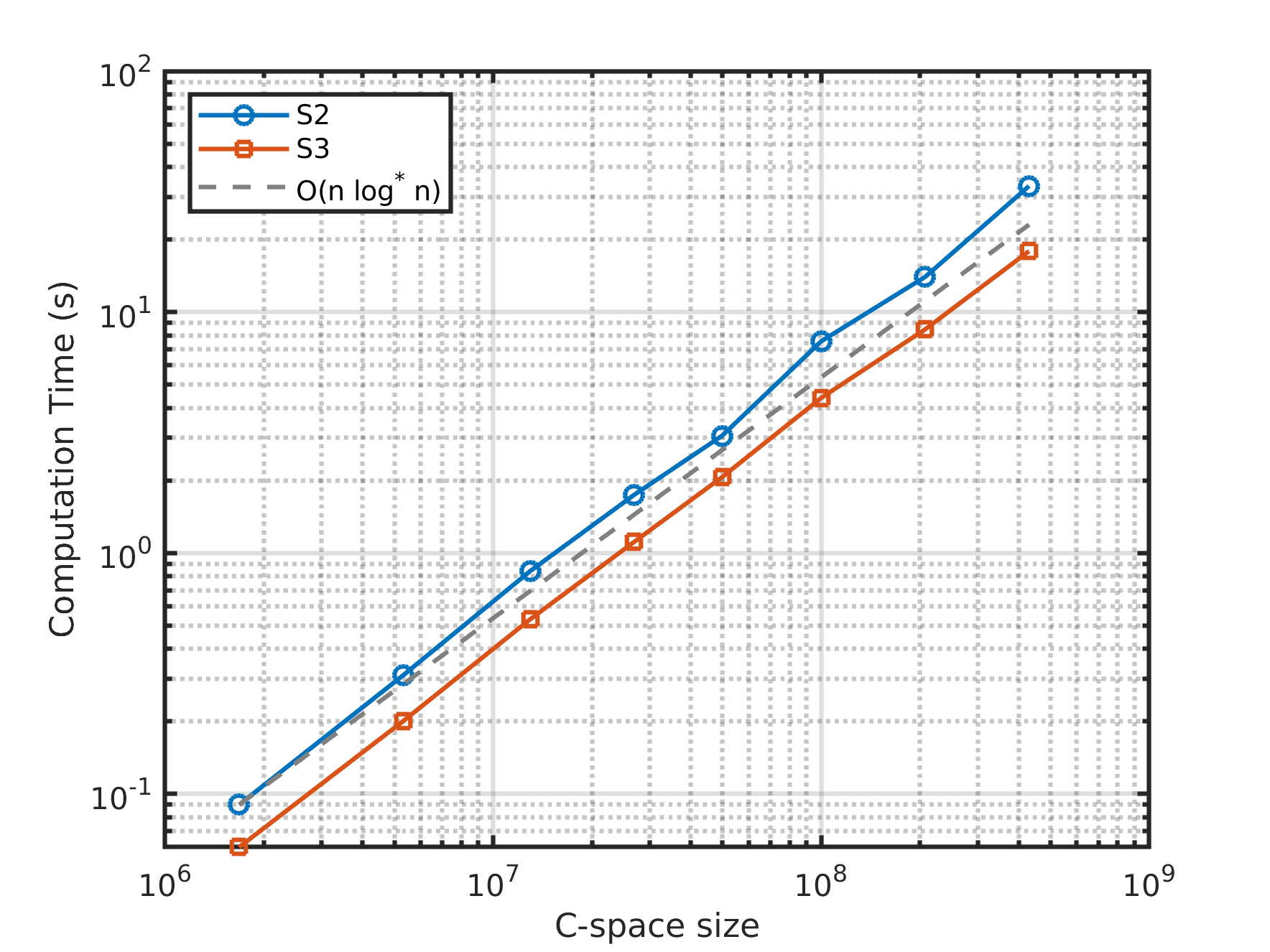}\label{fig:astar1}
                    \caption{Segmentation times for varying configuration-space resolutions. The observed trends are consistent with the theoretical $O(n \log^* n)$ complexity of the union–find algorithm.}
                     \label{fig:segmentation_time}
\end{figure}

We further evaluate the segmentation performance across scenarios $S_2$ and $S_3$ for varying C-space resolutions corresponding to $\C{CB}$ sizes of $36^4$, $48^4$, $60^4$, $72^4$, $84^4$, $100^4$, $120^4$, and $144^4$. The corresponding segmentation times are plotted in Fig.~\ref{fig:segmentation_time}. As observed from the figure, the measured computation times closely follow the expected $O(n \log^* n)$ growth trend, thereby validating the near-linear complexity of the segmentation algorithm.

\subsection{Choosing $ns$ and $d$}
We first examine the effect of fixing $d$. While lower values of $d$ resulted in similar overall runtime performance, higher values ($d > 5$) led to increased runtimes. This is because checking $d$ neighboring samples for collisions requires additional calls to the collision checker, thereby increasing computational cost. Based on our observations, $d = 5$ provides a practical balance.

Independently of $d$, smaller values of $ns$ ($ns < 100$) did not significantly impact performance for scenarios $S_1$-$S_4$, as the segmentation times are negligible. However, larger values of $ns$ could lead to longer runtimes simply due to the acquisition of more samples (and thereby calls to \texttt{CollisionCheck}) than necessary to establish infeasibility. In $S_5$, where segmentation time constitutes a major portion of the total runtime, setting $ns < 100$ could result in multiple iterations (as opposed to the single iteration currently required for $S_5$, as shown in the sixth column of Table~\ref{tab:statistics}), increasing runtime. The same reasoning extends to $S_6$, for which $ns < 300$ leads to more than one iteration.

To provide a quantitative assessment, we fix $ns = 100$ and vary $d \in \{0, 3, 5, 8,$ $ 15, 25\}$ for scenarios $S_3$ and $S_4$, analyzing the resulting computation times across three different C-space discretization densities. The results, summarized in Table~\ref{tab:d_vary}, reveal a consistent trend across both scenarios and all resolutions: the runtime initially decreases as $d$ increases from 0 to 5, indicating improved efficiency from incorporating limited local neighborhood information. Beyond this range ($d > 8$), the runtime increases again due to the additional collision-checking overhead associated with larger neighborhoods. 

\begin{figure}[]
% \captionsetup{font={color=revisioncolor}}
\centering
  \subfloat[]{\includegraphics[trim=0 0 0 10,clip,scale=0.4]{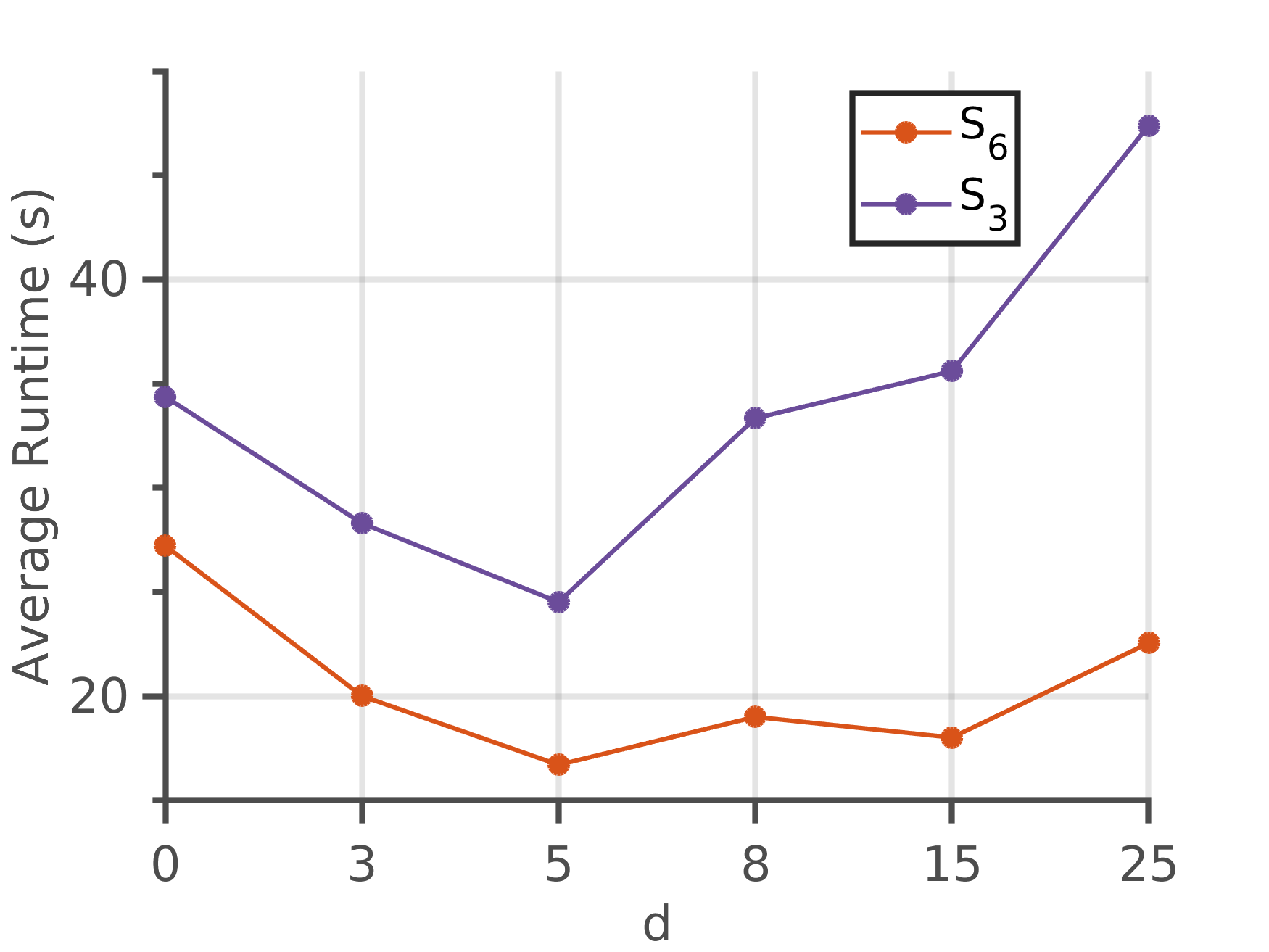}\label{fig:s3_s6_changingD}}\hspace{0.1cm}
      \subfloat[]{\includegraphics[trim=1 1 1 10,clip,scale=0.4]{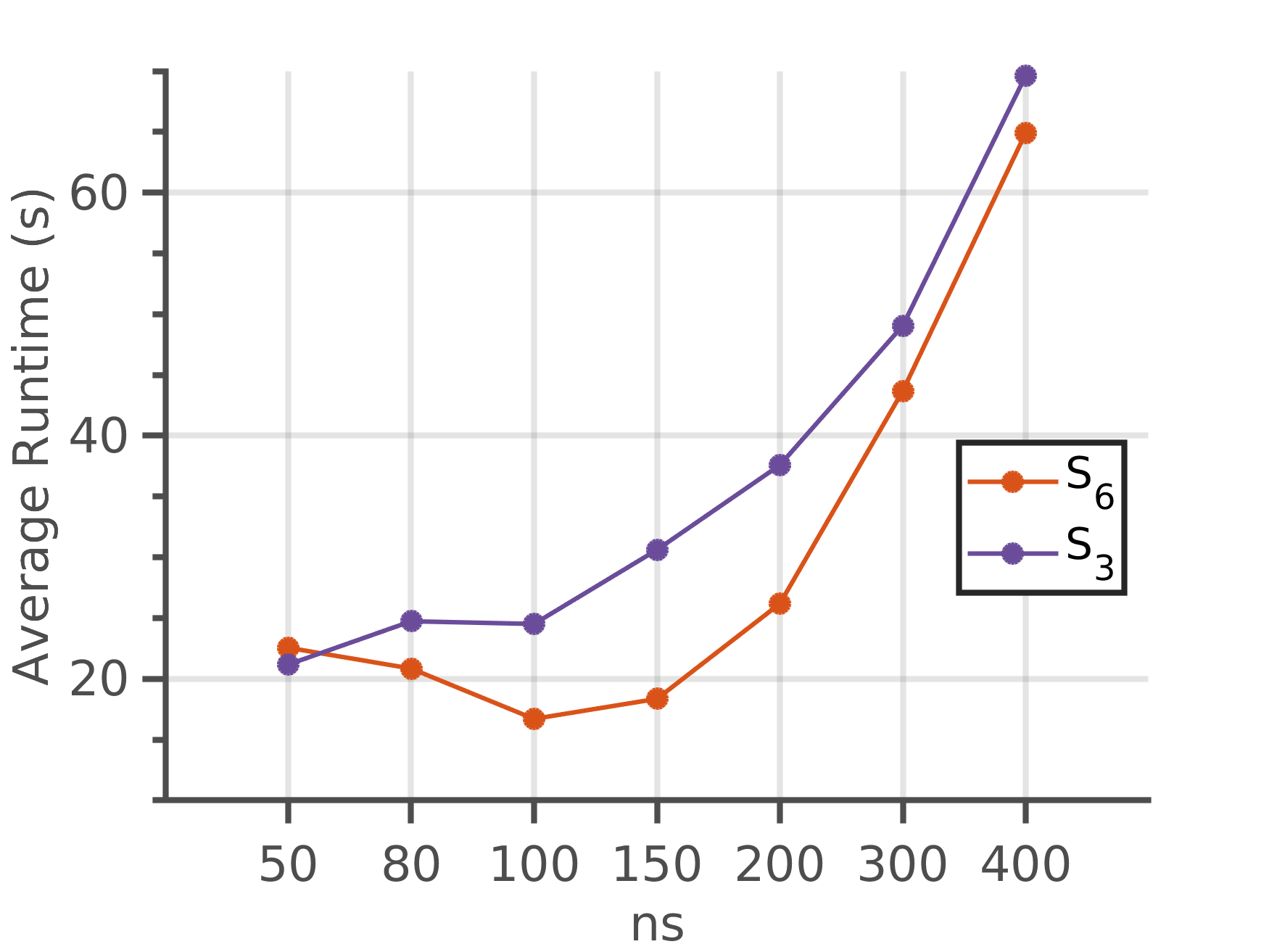}\label{fig:s3_s6_changingNS}}
                    \caption{Runtime as a function of $d$ (a) and $ns$ (b) for $S_3$ and $S_6$, with $\C{CB}$ resolution fixed at $36$. Similar trends are observed across varying parameter values for both scenarios.}
                     \label{fig:param_changing}
\end{figure}

For instance, in $\C{CB}$ of size $36^4$ for $S_3$, the runtime decreases from 34.37 s ($d=0$) to 24.52 s ($d=5$), but rises to 47.38 s at $d=25$. A similar pattern is observed for $S_4$, where the runtime reduces from 47.54 s ($d=0$) to 28.10 s ($d=5$), before increasing again to 48.72 s at $d=25$. This trend is consistent at higher resolutions as well—for example, in the $\C{CB}$ of size $72^4$, $S_3$ shows a sharp drop from 219.51 s to 107.27 s between $d=0$ and $d=5$, while $S_4$ drops from 247.61 s to 136.01 s over the same range. These results collectively demonstrate that $d = 5$ consistently achieves the best computational trade-off across both scenarios. We also perform the same analysis for the 5-DOF scenario $S_6$ at a resolution of $36^\circ$. A trend similar to that observed in Table~\ref{tab:d_vary} is obtained. To make this explicit, we compare the runtimes with $S_3$, as illustrated in Fig.~\ref{fig:s3_s6_changingD}.
\begin{table}[t]
%\arrayrulecolor{revisioncolor}
% \captionsetup{font={color=revisioncolor}}
% \color{revisioncolor}
\centering
\scalebox{0.69}{
\begin{tabular}{c|cccccc}
\hline
% \rule{0pt}{1.05\normalbaselineskip}
\multirow{2}{*}{C-space size} & \multicolumn{6}{c}{Average runtime (s) for \textbf{Scenario $\textbf{S}_{\textbf{3}}$}} \\ 
\cline{2-7}
 & $d=0$ & $d=3$ & $d=5$ & $d=8$ & $d=15$ & $d=25$ \\
\hline
\hline
\rule{0pt}{1.05\normalbaselineskip}
36$\times$36$\times$36$\times$36 & 34.37$\pm$13.20 & 28.30$\pm$6.88 & 24.52$\pm$9.18 & 33.34$\pm$8.80 & 35.60$\pm$15.61 & 47.38$\pm$18.91 \\
48$\times$48$\times$48$\times$48 & 51.31$\pm$13.39 & 41.34$\pm$16.14 & 42.93$\pm$12.42 & 36.78$\pm$9.65 & 62.51$\pm$24.14 & 65.91$\pm$17.11 \\
72$\times$72$\times$72$\times$72 & 219.51$\pm$26.75 & 111.29$\pm$20.76 & 107.27$\pm$19.99 & 120.25$\pm$25.37 & 123.83$\pm$30.56 & 149.25$\pm$33.48 \\
\hline
\multirow{2}{*}{C-space size} & \multicolumn{6}{c}{Average runtime (s) for \textbf{Scenario $\textbf{S}_{\textbf{4}}$}} \\ 
\cline{2-7}
 & $d=0$ & $d=3$ & $d=5$ & $d=8$ & $d=15$ & $d=25$ \\
\hline
\hline
\rule{0pt}{1.05\normalbaselineskip}
36$\times$36$\times$36$\times$36 & 47.54$\pm$14.52 & 30.98$\pm$13.79 & 28.10$\pm$12.87 & 38.89$\pm$15.33 & 40.14$\pm$13.95 & 48.72$\pm$11.36 \\
48$\times$48$\times$48$\times$48 & 102.84$\pm$47.90 & 58.86.34$\pm$19.88 & 56.31$\pm$25.24 & 68.72$\pm$27.68 & 71.14$\pm$38.36 & 90.57$\pm$40.51 \\
72$\times$72$\times$72$\times$72 & 247.61$\pm$103.49 & 146.93$\pm$47.25 & 136.01$\pm$49.23 & 122.25$\pm$53.18 & 131.17$\pm$31.42 & 149.83$\pm$42.69 \\
\hline
\end{tabular}}
\caption{Average runtime over 30 runs for scenarios $S_3$ and $S_4$ under varying values of the neighborhood parameter $d$. The number of samples $ns$ is fixed at 100, and results are reported for three $\C{CB}$ sizes.}
\label{tab:d_vary}
\end{table}

To examine the effect of the sampling density on computational performance, we fix the neighborhood parameter at $d=5$ and vary the number of samples $ns \in \{50, 80, 100, 150, 200, 300, 400\}$ for scenarios $S_3$ and $S_4$. Table~\ref{tab:ns_vary} reports the corresponding average runtimes over 30 runs for three different C-space discretization densities. A consistent trend is observed across both scenarios: the runtime increases gradually with $ns$, reflecting the additional computational cost of generating and evaluating larger sample sets. However, the rate of increase is sublinear for moderate $ns$ values, indicating that the cost of sampling amortizes well up to about $ns=150$.

For instance, in $\C{CB}$ of size $36^4$ for $S_3$, the runtime increases moderately from 21.18 s ($ns=50$) to 30.62 s ($ns=150$), but rises sharply to 69.63 s at $ns=400$. A similar pattern is evident for $S_4$, where the runtime grows from 27.97 s at $ns=50$ to 34.09 s at $ns=150$, before reaching 103.42 s at $ns=400$. Larger C-space resolutions exhibit the same qualitative behavior, with proportionally higher runtimes due to the increased collision-checking cost per configuration.  Overall, these results suggest that $ns$ values between 80 and 150 offer a favorable trade-off between computational efficiency and sampling coverage. 

An interesting case is observed for $ns = 300$ and $ns = 400$ in scenario $S_3$, where both C-space resolutions result in similar runtimes. For $ns < 300$, Algorithm~\ref{algo:MPI} requires multiple iterations (see Table~\ref{tab:statistics}, row 3, column 6). In contrast, for $ns \geq 300$, Algorithm~\ref{algo:MPI} converges in a single iteration. As a result, the elimination of these repeated iteration overheads offsets the additional cost associated with the higher C-space resolution. Consequently, the only remaining difference is the segmentation time, which is negligible for the considered C-space resolutions (see Table~\ref{tab:statistics}, row 3, column 7). The same reasoning also applies to scenario $S_4$, for $ns=400$, where the C-space resolutions $36^4$ and $48^4$ exhibit similar runtimes.

We also examine the effect of varying $ns$ for the the 5-DOF scene $S_6$. A similar trend in computation time is observed as in the 4-DOF scenarios, as shown in Fig.~\ref{fig:s3_s6_changingNS}.

\begin{table}[]
% \captionsetup{font={color=revisioncolor}}
%\arrayrulecolor{revisioncolor}
% \color{revisioncolor}
\centering
\setlength{\tabcolsep}{3pt}
\scalebox{0.66}{
\begin{tabular}{c|ccccccc}
\hline
\multirow{2}{*}{C-space size} & \multicolumn{7}{c}{Average runtime (s) for \textbf{Scenario $\textbf{S}_{\textbf{3}}$}} \\ 
\cline{2-8}
 & $ns=50$ & $ns=80$ & $ns=100$ & $ns=150$ & $ns=200$ & $ns=300$ & $ns=400$ \\
\hline
\hline
\rule{0pt}{1.05\normalbaselineskip}
36$\times$36$\times$36$\times$36 & 21.18$\pm$6.03 & 24.74$\pm$8.02 & 24.52$\pm$9.18 & 30.62$\pm$12.61 & 37.59$\pm$14.00 & 49.06$\pm$2.09 & 69.63$\pm$3.07 \\
48$\times$48$\times$48$\times$48 & 34.16$\pm$13.64 & 36.10$\pm$11.88 & 42.93$\pm$12.42 & 38.92$\pm$12.47 & 43.99$\pm$17.48 & 49.88$\pm$10.83 & 66.19$\pm$1.79 \\
72$\times$72$\times$72$\times$72 & 108.83$\pm$23.10 & 111.64$\pm$29.98 & 107.27$\pm$19.99 & 122.99$\pm$21.87 & 123.70$\pm$27.04 & 133.83$\pm$35.44 & 145.61$\pm$41.36 \\
\hline
\multirow{2}{*}{C-space size} & \multicolumn{7}{c}{Average runtime (s) for \textbf{Scenario $\textbf{S}_{\textbf{4}}$}} \\ 
\cline{2-8}
 & $ns=50$ & $ns=80$ & $ns=100$ & $ns=150$ & $ns=200$ & $ns=300$ & $ns=400$ \\
\hline
\hline
\rule{0pt}{1.05\normalbaselineskip}
36$\times$36$\times$36$\times$36 & 27.97$\pm$7.70 & 29.69$\pm$10.56 & 28.10$\pm$12.87 & 34.09$\pm$8.25 & 49.17$\pm$11.25 & 74.66$\pm$2.79 & 103.42$\pm$5.14 \\
48$\times$48$\times$48$\times$48 & 54.53$\pm$26.22 & 70.27$\pm$41.06 & 56.31$\pm$25.24 & 61.24$\pm$20.17 & 65.59$\pm$30.64 & 96.08$\pm$38.90 & 106.97$\pm$29.51 \\
72$\times$72$\times$72$\times$72 & 127.87$\pm$43.29 & 122.83$\pm$51.63 & 136.01$\pm$49.23 & 132.97$\pm$45.72 & 151.96$\pm$50.85 & 135.72$\pm$50.87 & 162.05$\pm$54.81 \\
\hline
\end{tabular}}
\caption{Average runtime over 30 runs for scenarios $S_3$ and $S_4$ under varying numbers of samples $ns$, with $d=5$ fixed for all cases. Increasing $ns$ generally leads to higher runtimes, reflecting the added sampling and evaluation cost. The growth is sublinear for moderate $ns$ values ($\leq150$), indicating efficient amortization of sampling overhead. Beyond this range, the runtime increases sharply, suggesting diminishing computational returns for larger sample sets.}
\label{tab:ns_vary}
\end{table}

\begin{figure}[]
\centering
  \subfloat[]{\includegraphics[trim=115 150 34 40,clip,scale=0.52]{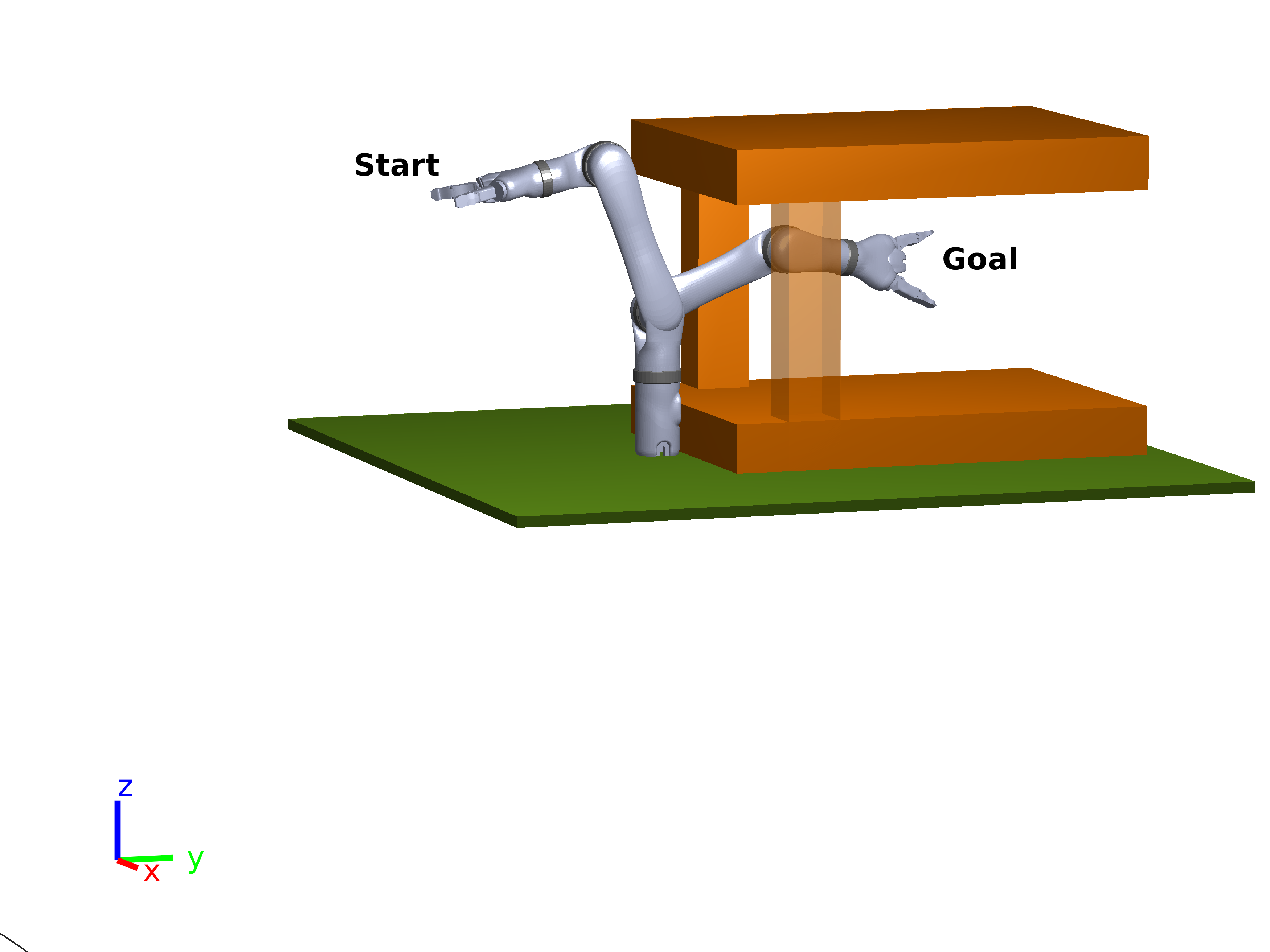}\label{fig:dis1}}\hspace{0.1cm}
      \subfloat[]{\includegraphics[trim=50 120 50 40,clip,scale=0.45]{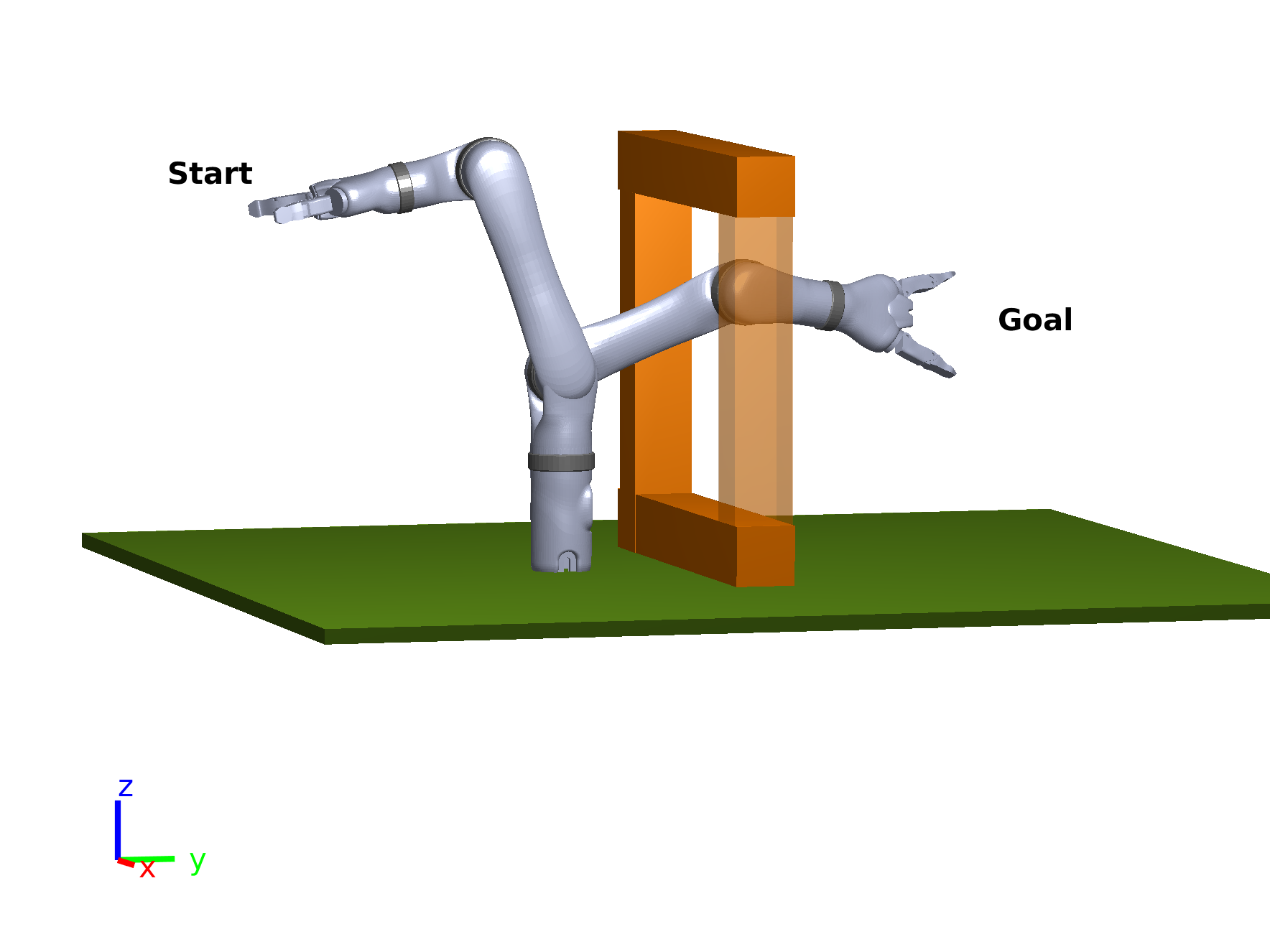}\label{fig:dis2}}
                    \caption{(a) Modified version of $S_3$ with an obstacle volume 2.4 times larger than in $S_3$. The increased obstacle size results in a broader C-obstacle region, facilitating sampling but potentially leading to unnecessary computations. (b) Obstacle volume reduced to $4/11$ of that in $S_3$. The thinner obstacle makes sampling from the C-obstacle more challenging, leading to higher runtime.}
                     \label{fig:dis}
\end{figure}
 \begin{figure}[t]
 \centering
  \subfloat[]{\includegraphics[trim=0 0 12 5,clip,scale=0.4]{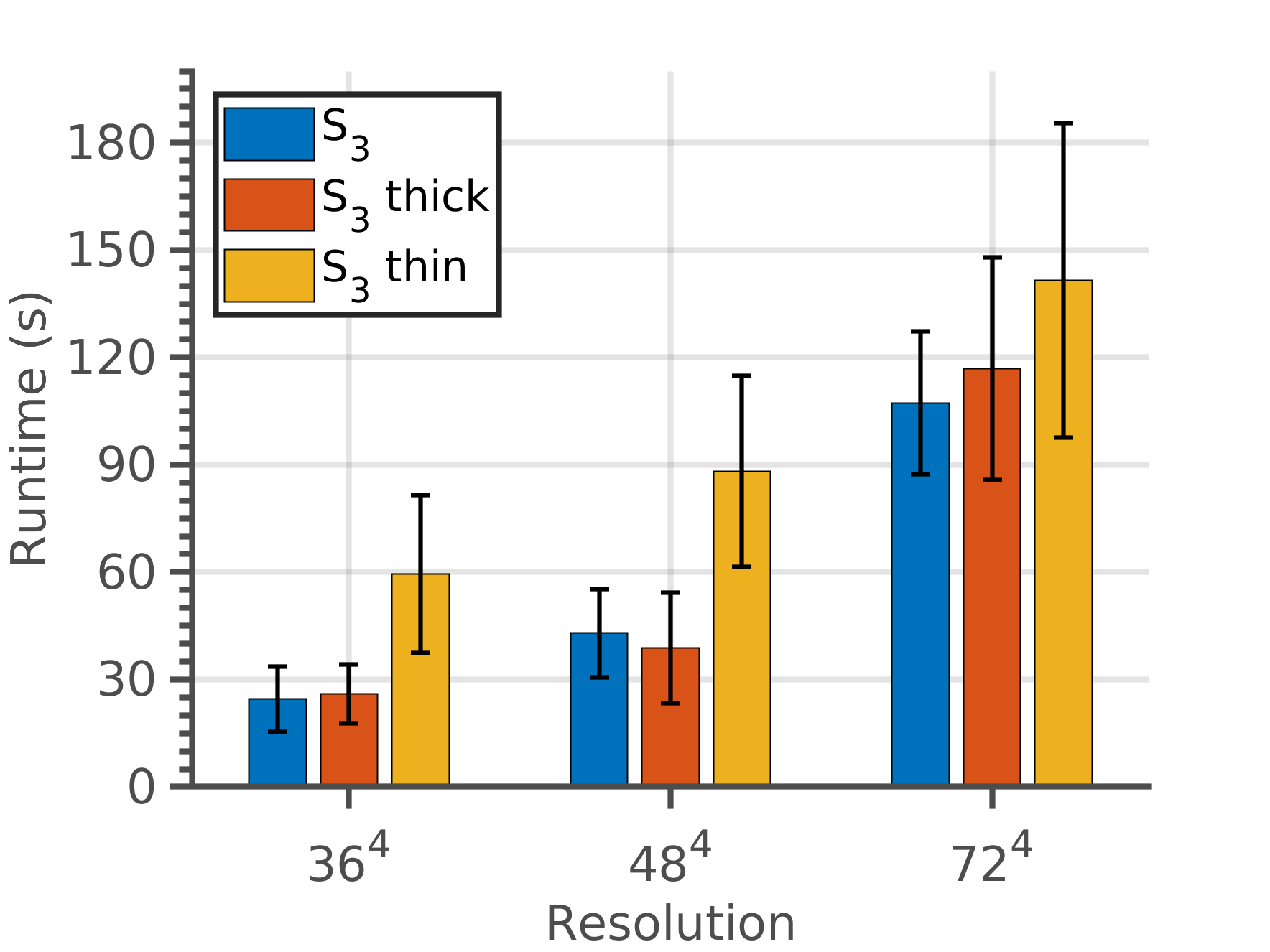}\label{fig:sd}}
      \subfloat[]{\includegraphics[trim=0 0 12 5,clip,scale=0.4]{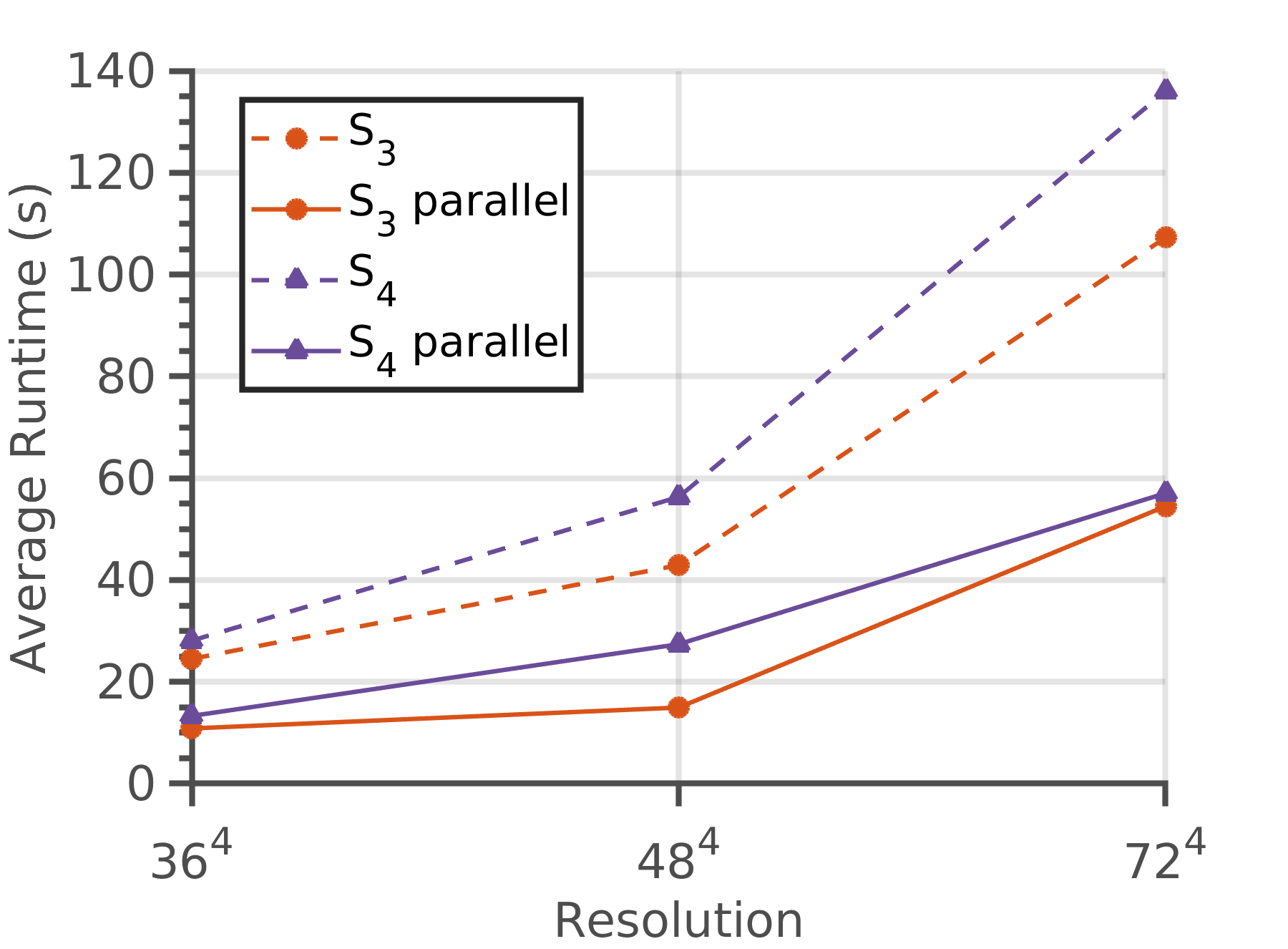}\label{fig:np}}
     \caption{(a) Runtime comparison for $S_3$, the thick-obstacle case ($2.4 \times S_3$ obstacle volume), and the thin-obstacle case ($4/11 \times S_3$ obstacle volume). Runtime for the thick-obstacle case remains comparable to $S_3$, whereas the thin-obstacle case exhibits a higher runtime with increased standard deviation. (b) Comparison of average computation time for $S_3$ and $S_4$ in both sequential and parallel implementations.}
  \label{fig:discussion}
\end{figure}

\subsection{Obstacle Volume}
As discussed in Section~\ref{sec:approach}, the proposed method is based on the principle that it is not necessary to sample all points in the C-obstacle, but only those required to establish infeasibility. The primary computational challenge lies in determining whether a sampled configuration belongs to the C-free or C-obstacle region, which necessitates executing the \texttt{CollisionCheck} subroutine.  

Thin obstacles in the workspace (those occupying less volume) correspond to narrow C-obstacle regions in the C-space, making it more challenging to sample from the C-obstacle. This is because a large proportion of random samples may correspond to collision-free configurations. Conversely, thick obstacles (those with larger volume) result in broader C-obstacle regions, making sampling easier. However, in such cases, an excessive number of configurations may be sampled, potentially exceeding what is necessary to establish infeasibility, thereby leading to increased runtime.  

To analyze this effect, Fig.~\ref{fig:dis} presents two modified versions of $S_3$ (Fig.~\ref{fig:3D1}). In Fig.~\ref{fig:dis1}, the obstacle volume is 2.4 times that of $S_3$, whereas in Fig.~\ref{fig:dis2}, it is reduced to $4/11$ of the volume of $S_3$. The runtime comparison (average across 30 trials) for all the scenarios (including $S_3$) is provided in Fig.~\ref{fig:sd}. For the thick-obstacle case, the runtime remains comparable to $S_3$. However, for the thin-obstacle case, as expected, the runtime increases, also exhibiting a larger standard deviation.  

Currently, no specialized sampling technique is employed; instead, we randomly sample without replacement from the $\C{CB}$ bitmap to check for occupancy. A more engineered approach to sampling from C-obstacle could significantly mitigate the above bottleneck by avoiding the selection of numerous C-free elements, thereby reducing unnecessary calls to the \texttt{CollisionCheck} subroutine.
\subsection{Complexity Analysis}
A rigorous theoretical characterization of the computational complexity of the proposed approach is nontrivial and lies beyond the scope of this study. To gain empirical insight, we analyzed the scaling behavior of computation time for scenarios $S_2$ and $S_3$ by varying the C-space discretization with grid sizes of $36^4$, $48^4$, $60^4$, $72^4$, $84^4$, $100^4$, $120^4$, and $144^4$. The measured computation times were fitted against several functional models, including linear, quadratic, and cubic polynomials, as well as power, exponential, and $x \log x$ relationships.

\begin{figure}[]
% \captionsetup{font={color=revisioncolor}}
\centering
\subfloat[]{\includegraphics[width=0.52\textwidth]{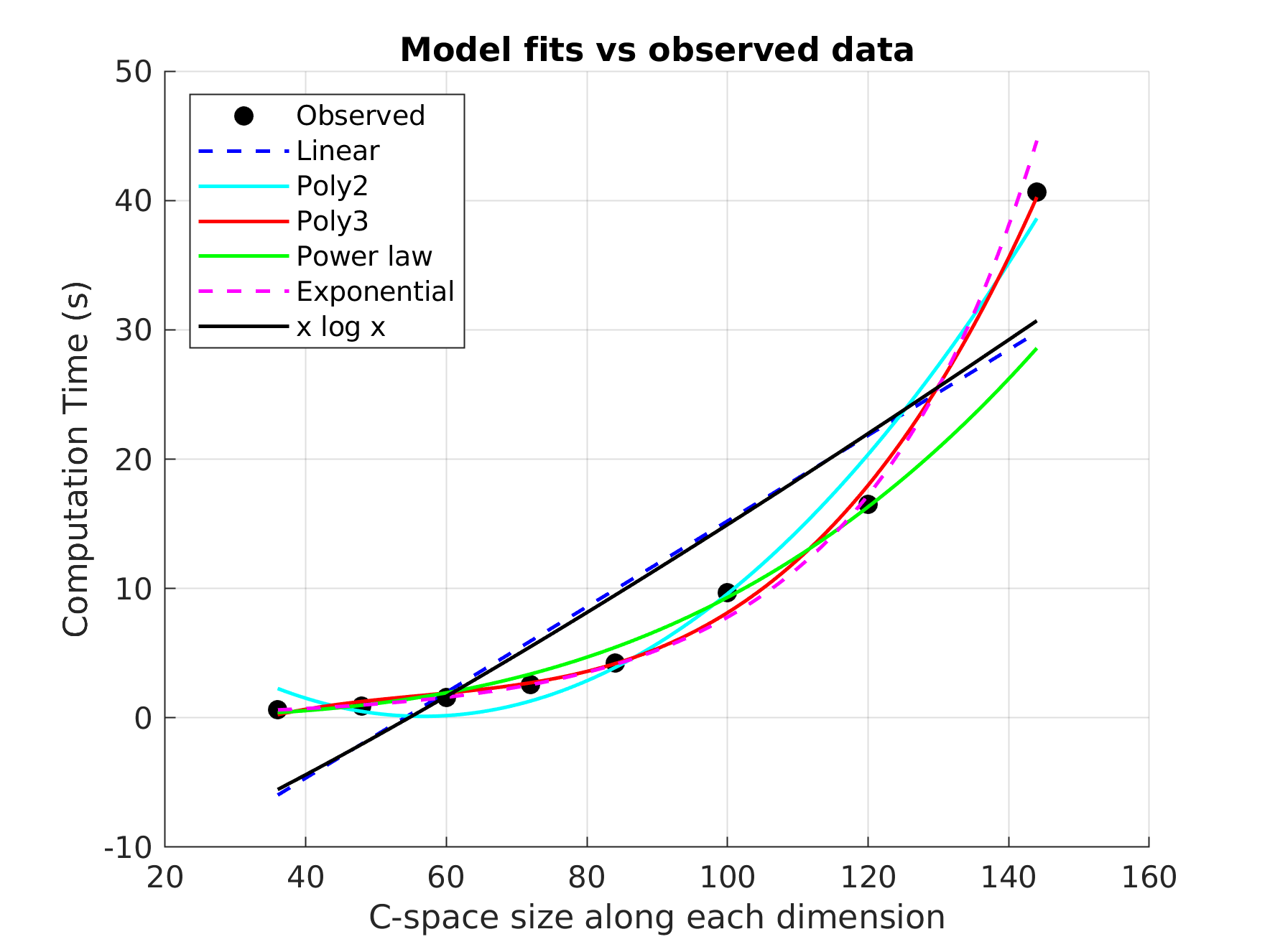}\label{fig:s2time}}
\subfloat[]{\includegraphics[width=0.52\textwidth]{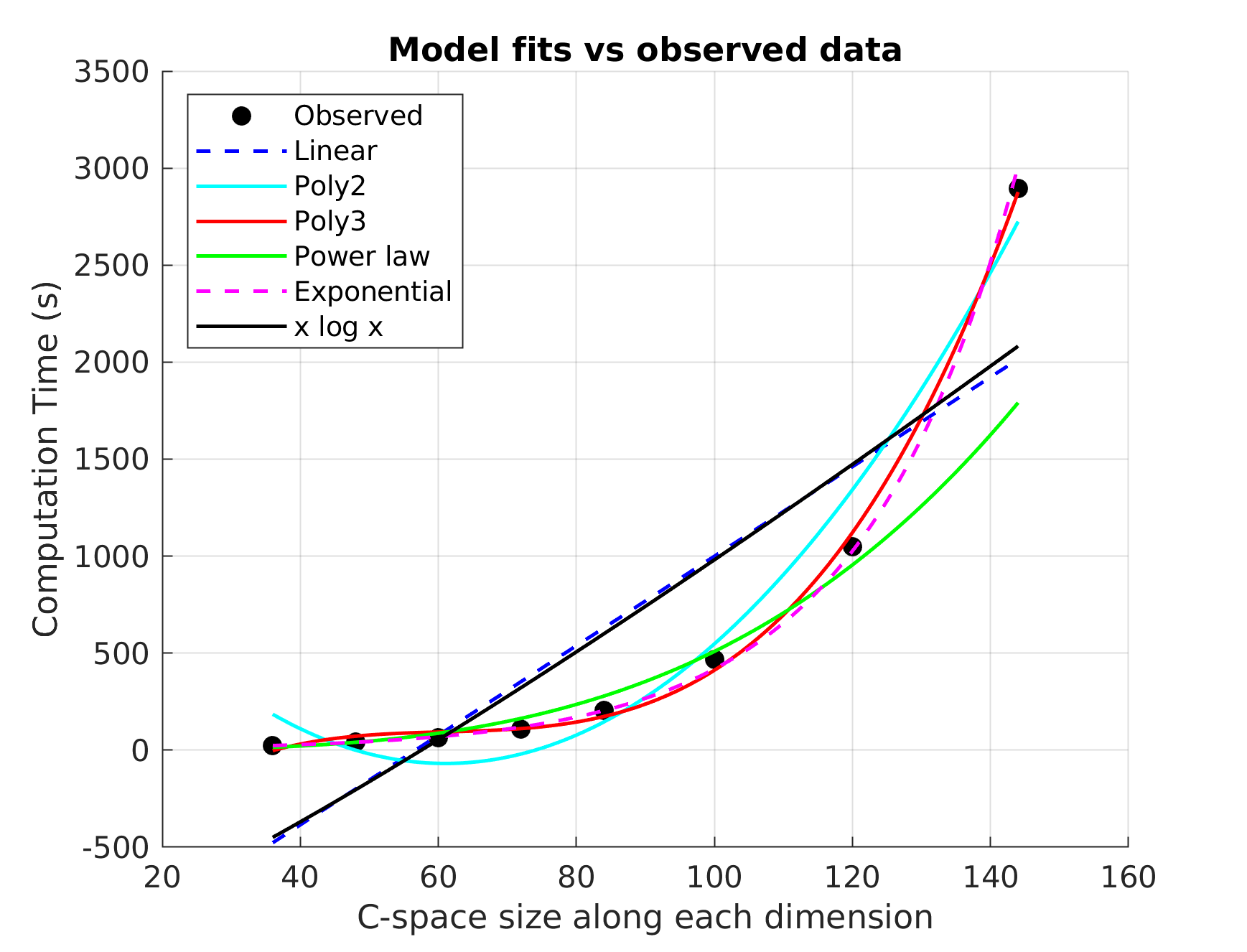}\label{fig:s3time}}
\caption{Empirical complexity analysis for scenarios (a) $S_2$ and (b) $S_3$. Computation times were fitted using multiple functional models, with a third-order polynomial in the C-space size per dimension providing the best fit.}
\label{fig:complexity}
\end{figure}

Model fitting was evaluated using both the coefficient of determination and mean squared error metrics. Across both scenarios, a third-order polynomial in the C-space size per dimension yielded the best fit, indicating a cubic dependence of computation time on the discretization density of the C-space. The fitted curves and experimental data are presented in Fig.~\ref{fig:complexity}.
\subsection{Runtime Reduction} 
The \texttt{CollisionCheck} subroutine, particularly for scenarios $S_3$ and $S_4$, remains the most computationally expensive component. Implementing faster collision-checking techniques~\cite{das2020TRO, montaut2022RSS} can significantly accelerate the process. Another approach to enhancing computational efficiency is to parallelize Algorithm~\ref{algo:samplecobstacle}. However, in its current form, Algorithm~\ref{algo:samplecobstacle} is not embarrassingly parallel, as the \texttt{RandomSample} subroutine may generate identical samples across parallel tasks. If this limitation is disregarded, the \textit{for loop} can be naively parallelized.  

To this end, we rerun $S_3$ and $S_4$ by parallelizing Algorithm~\ref{algo:samplecobstacle} using MATLAB’s Parallel Computing Toolbox, enabling execution across four cores. The average runtimes are compared in Fig.~\ref{fig:np}, where it can be observed that the computation time is reduced by about 2.4 times for both $S_3$ and $S_4$. The computations were conducted on a Lenovo ThinkPad laptop equipped with an Intel{\small\textregistered} Core i7-10510U CPU$@$1.80GHz$\times$8 and 16GB RAM running Ubuntu 18.04 LTS. Leveraging the parallelism of modern high-performance CPUs and GPUs can further improve computational efficiency.  
\begin{table}[t]
\begin{center}
\begin{tabular}{c c c c c c} 
\hline
\rule{0pt}{1.005\normalbaselineskip}
Scenarios & DOF & Resolution & $ns$ & $d$ & Total time (s) \\
\hline
\hline
\rule{0pt}{1.005\normalbaselineskip}
 \multirow{3}{*}{$S_7$ (Fig.~\ref{fig:6dof})} & \multirow{3}{*}{6} & 36$\times$36$\times$36 & \multirow{3}{*}{100} & \multirow{3}{*}{5}& 5.08$\pm$2.03 \\
 &  & 48$\times$48$\times$48 & & & 8.02$\pm$1.69\\
 &  & 72$\times$72$\times$72 & & & 14.08$\pm$2.71\\
 \hline
   \rule{0pt}{1.005\normalbaselineskip}
 \multirow{3}{*}{$S_8$ (Fig.~\ref{fig:7dof})} & \multirow{3}{*}{7} & 36$\times$36$\times$36 & \multirow{3}{*}{100} & \multirow{3}{*}{5} & 14.51$\pm$4.57 \\
  &  & 48$\times$48$\times$48 & & & 20.10$\pm$5.45\\
 &  & 72$\times$72$\times$72 & & & 45.83$\pm$23.81\\
  \hline
  \end{tabular}
\end{center}
\caption{Runtime (average across 30 trials) comparison for infeasibility verification in the 6-DOF and 7-DOF robot arm experiments demonstrates the efficiency gains achieved by reducing the C-space dimensionality. Infeasibility is proven within a 3D C-space for both $S_7$ and $S_8$.}
\label{tab:statistics2}
\end{table}
\subsection{Feasible Plan Experiments}
\label{subsec:feasible_infeasible}
\begin{algorithm}
% \color{revisioncolor}
\caption{Parallel Motion Planning with Infeasibility Detection}
\label{algo:planning_infeasible}
\begin{algorithmic}[1]
\Require{$\C{R}$, $\C{O}$, $q_s$, $q_g$}
\Ensure{Feasible path $\mathcal{P}$ or infeasibility certificate}
\State{Launch $\C{F}_1 \leftarrow \texttt{MotionPlanning}(\C{R}, \C{O}, q_s, q_g)$} 
\LineComment \textcolor{mygray}{Async worker for motion planning algorithm.}
\State{Launch $\C{F}_2 \leftarrow \texttt{MotionPlanningInfeasibilityDetection}(\C{R}, \C{O}, q_s, q_g)$} \LineComment \textcolor{mygray}{Async worker for infeasibility detection.}
\State{ $(i^*, \C{r}) \leftarrow \texttt{FetchNext}(\{\C{F}_1, \C{F}_2\})$} 
\LineComment \textcolor{mygray}{Block until one worker completes.}
\State{\textsc{Cancel}$(\C{F}_1)$, \textsc{Cancel}$(\C{F}_2)$} 
\LineComment \textcolor{mygray}{Terminate remaining worker.}
\If{$i^* = 1$}
    \State{\Return feasible path $\C{P}$}
\Else
    \State {\Return infeasibility certificate}
\EndIf
\Statex
\Statex \textbf{Subroutine} \texttt{MotionPlanning}
\LineComment \textcolor{mygray}{Motion Planning Algorithm.}
\Statex \Return \text{$\C{P}$}
\Statex
\Statex \textbf{Subroutine} \texttt{MotionPlanningInfeasibilityDetection}
\LineComment \textcolor{mygray}{Algorithm~\ref{algo:MPI}.}
\Statex \Return infeasibility certificate
\end{algorithmic}
\end{algorithm}
Though the primary goal of this work is to certify infeasibility, the approach can be naturally extended to run a motion planner in parallel with Algorithm~\ref{algo:MPI}. In this setting, if a feasible plan exists, the motion planner returns a solution; otherwise, if the problem is infeasible, Algorithm~\ref{algo:MPI} provides a certificate of infeasibility. Both the motion planner and the infeasibility detection algorithm are executed concurrently on independent workers. The main thread waits at a synchronization point and resumes as soon as the first worker completes. Upon completion, the remaining worker is immediately terminated. This parallel formulation ensures that feasibility is established as quickly as possible when a solution exists, while still guaranteeing certification of infeasibility otherwise. Algorithm~\ref{algo:planning_infeasible} summarizes this parallel strategy.

We modify scenarios $S_3$, $S_4$, and $S_6$ to ensure the existence of a feasible path. For $S_3$ and $S_6$, the frame is translated away from the robot base to make reaching inside the frame feasible. The magnitude of translation differs across the two cases, resulting in a narrower passage in $S_3$ compared to $S_6$. For $S_4$, the red block on the right side of the scene is removed, thereby creating a narrow passage. The resulting planning times for obtaining a feasible path are reported in Table~\ref{tab:feasible}. Path planning is performed using RRT-Connect.
\begin{table}[t]
% \arrayrulecolor{revisioncolor}
% \color{revisioncolor}
\begin{center}
\begin{tabular}{c c c c c c} 
\hline
Scenarios & Planning time (s) \\
\hline
\hline
\rule{0pt}{1.005\normalbaselineskip}
 $S_6$ (Fig.~\ref{fig:3D_5dof}) & 2.69\\
 $S_3$ (Fig.~\ref{fig:3D1}) & 39.51\\
$S_4$ (Fig.~\ref{fig:3D2}) & 119.71\\
  \hline
  \end{tabular}
\end{center}
\caption{Average runtime for feasible plan experiments.}
\label{tab:feasible}
\end{table}
\subsection{Scaling to Higher Dimensions}
The proposed approach is effective for up to 5-DOF C-spaces. Higher-dimensional C-spaces introduce complexity in both space (for representing $\C{CB}$) and time (for segmentation). However, we make the following observation: in many motion planning problems involving a manipulator, infeasibility is often attributed to a specific link, typically one of the earlier links in the kinematic chain. Consequently, infeasibility proofs can often be formulated in lower-dimensional spaces while remaining valid in higher-dimensional spaces--- an approach grounded in quotient space topology~\cite{orthey2018IROS, orthey2019ISRR}.

\begin{figure}[t!]
\centering
  \subfloat[]{\includegraphics[trim=0 60 20 30,clip,scale=0.42]{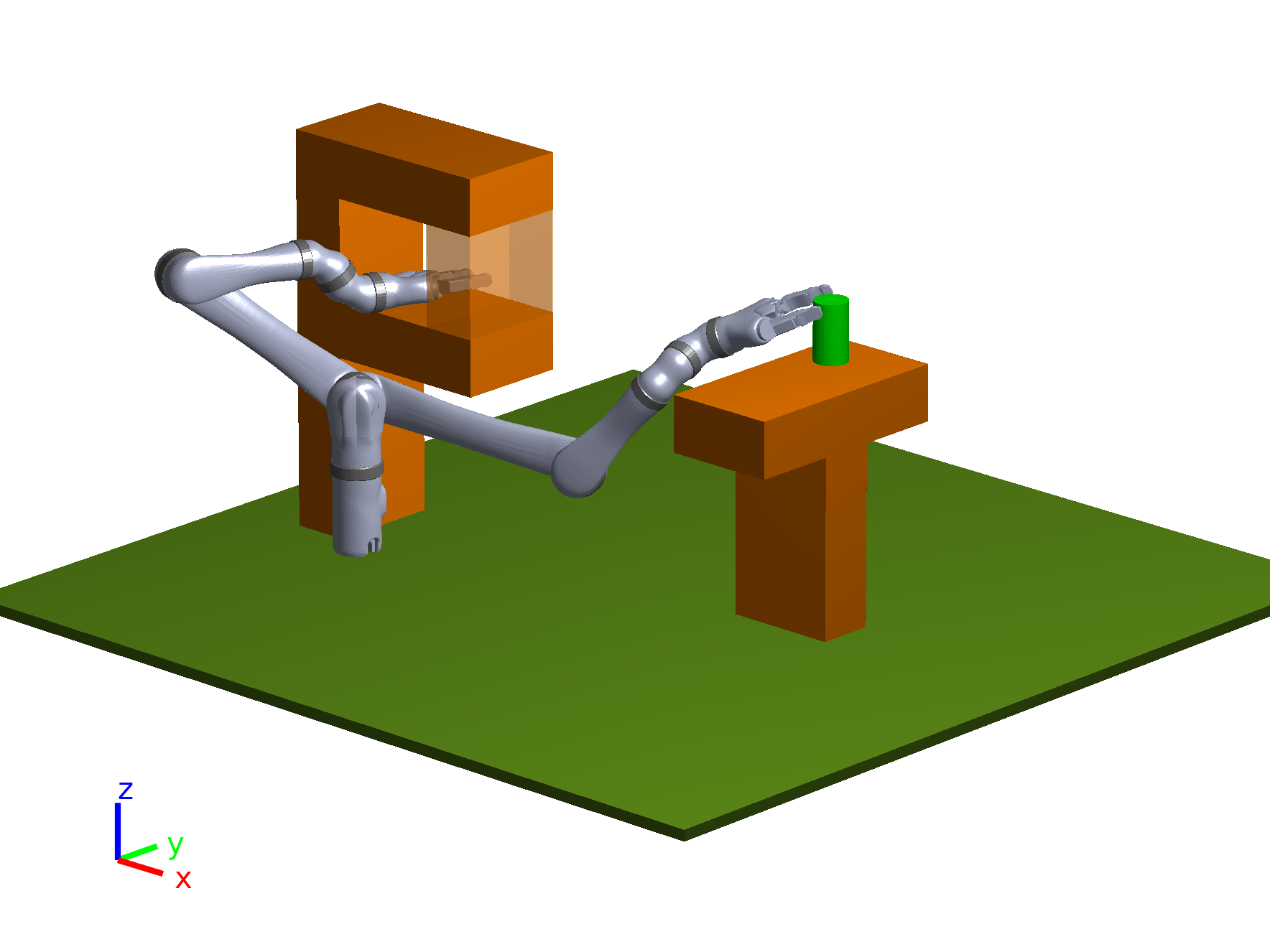}\label{fig:6dof}}\hspace{0.1cm}
      \subfloat[]{\includegraphics[trim=80 80 20 50,clip,scale=0.51]{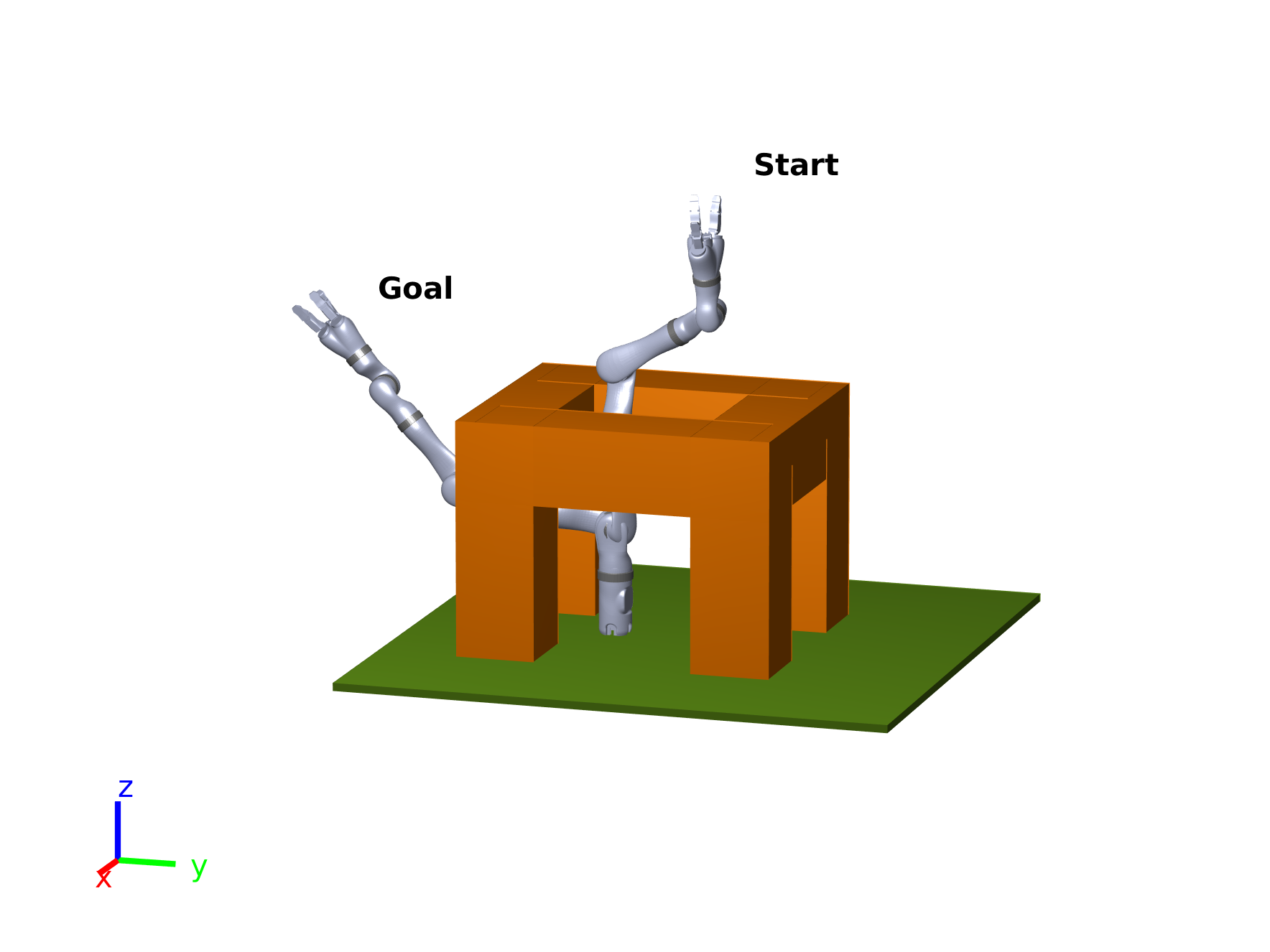}\label{fig:7dof}}
                    \caption{(a) A 6-DOF robot arm that needs to pick up the green object and placing it inside the shelf. (b) A 7-DOF robot arm trying to reach outside the table from the inside.}
  \label{fig:higherd}
\end{figure}

To illustrate this, we conduct experiments on two scenarios: (a) $S_7$, where a 6-DOF robot arm is tasked with picking up a green object and placing it inside a shelf (see Fig.~\ref{fig:6dof}), and (b) $S_8$, where a 7-DOF robot arm attempts to reach outside the table from the inside (Fig.~\ref{fig:7dof}). In both scenarios, infeasibility is caused by the third link, allowing us to consider a reduced 3-DOF C-space to establish infeasibility. The runtime results are presented in Table~\ref{tab:statistics2}, demonstrating that infeasibility for all three resolutions in both scenarios is proven in under 50 seconds.

While determining the appropriate reduced C-space remains a challenge, these examples suggest that many high-dimensional infeasibility problems can be addressed by proving infeasibility in a lower-dimensional C-space. Identifying such a minimal C-space dimension may require progressively building from lower dimensions. Nevertheless, pursuing these simplifications can significantly enhance the scalability of the proposed approach.  

%When checking for connectivity, the algorithm considers the full neighborhood in each dimension--- the level of connectivity in each dimension is passed as an input to the \textit{bwlabeln} function. For instance, in a 2-dimensional C-space, there can be a maximum of 8 neighbors, while in a 3-dimensional C-space, there can be up to 26 neighbors, and up to 80 in 4-dimensional spaces, and so on. Thus, for each sampled $q\in$ C-obstacle, the \texttt{SampleCobstacle} subroutine randomly selects $d$ neighbors from $q$'s full neighborhood.

%Plan infeasibility arises when $q_s$ and $q_g$ are situated in different connected components of the C-free space. Since we initialize all cells of the \(\mathcal{CB}\) as free regions, these cells need to be verified for occupancy. If the maximum resolution in any dimension of the C-space is \(N\), in the worst case, \(N^a\) cells have to be verified, where \(a \leq 5\). Thus, the overall computational complexity is \(O(N^a)\). However,  Consequently, for configuration spaces with up to 5 dimensions, the approach delivers reasonable performance in terms of computation time, detecting infeasibility in less than a minute.

%
%\section{Discussion}
%\label{sec:discussion}
%\input{discussion}
%
\section{Conclusion}
In this paper, we introduce a simple algorithm that is easy to implement for checking motion planning infeasibility. Our approach relies on the approximation of the C-space using a discretized bitmap. Initially, all cells of this bitmap are designated as C-free. Subsequently, our algorithm incrementally builds the bitmap by randomly sampling from the C-space and verifying the occupancy of each sampled cell. The constructed bitmap is then segmented into different connected components, which are separated by C-obstacles. Afterward, we query the connectivity of the start and goal cells. Through the incremental construction of the C-space, we only need to draw a sufficient number of samples to establish the partition between the start and goal configurations. The approach is validated through experiments involving robots with up 5-DOF. Additionally, we explore enhancements to accelerate the algorithm, presenting promising directions for future work. We also discuss scalability to higher dimensions and validate the approach through experiments on 6-DOF and 7-DOF robots. 
\section*{Acknowledgments}
The authors are grateful to the reviewers for their insightful comments and suggestions.

Work on this article by Antony Thomas is supported in part by the International Institute of Information Technology (IIIT) Hyderabad (Grant No. IIIT/R\&D Office/Seed-Grant/2025-26/002).
\bibliographystyle{elsarticle-num}
\bibliography{References}

@inproceedings{srivastava2014ICRA,
  title={Combined task and motion planning through an extensible planner-independent interface layer},
  author={Srivastava, Siddharth and Fang, Eugene and Riano, Lorenzo and Chitnis, Rohan and Russell, Stuart and Abbeel, Pieter},
  booktitle={Robotics and Automation (ICRA), IEEE International Conference on},
  pages={639--646},
  year={2014},
  organization={IEEE}
}

@book{sedgewick2004algorithms,
  title={{Algorithms in C}},
  author={Sedgewick, Robert},
  year={1998},
  publisher={3rd Ed., Addison-Wesley}
}

@article{kaelbling2013IJRR,
  title={Integrated task and motion planning in belief space},
  author={Kaelbling, Leslie Pack and Lozano-P{\'e}rez, Tom{\'a}s},
  journal={The International Journal of Robotics Research},
  volume={32},
  number={9-10},
  pages={1194--1227},
  year={2013},
  publisher={Sage Publications Sage UK: London, England}
}

@article{kavraki1996IEEE,
  title={Probabilistic roadmaps for path planning in high-dimensional configuration spaces},
  author={Kavraki, Lydia E and Svestka, Petr and Latombe, J-C and Overmars, Mark H},
  journal={IEEE Transactions on Robotics and Automation},
  volume={12},
  number={4},
  pages={566--580},
  year={1996},
  publisher={IEEE}
}

@inproceedings{kuffner2000ICRA,
  title={RRT-connect: An efficient approach to single-query path planning},
  author={Kuffner, James J and LaValle, Steven M},
  booktitle={Robotics and Automation, 2000. Proceedings. ICRA'00. IEEE International Conference on},
  volume={2},
  pages={995--1001},
  year={2000},
  organization={IEEE}
}

@article{garrett2018IJRR,
  title={{FFRob: Leveraging symbolic planning for efficient task and motion planning}},
  author={Garrett, Caelan Reed and Lozano-Perez, Tomas and Kaelbling, Leslie Pack},
  journal={The International Journal of Robotics Research},
  volume={37},
  number={1},
  pages={104--136},
  year={2018},
  publisher={SAGE Publications Sage UK: London, England}
}

@article{dantam2018IJRR,
    author = {Neil T. Dantam and Zachary K. Kingston and Swarat Chaudhuri and Lydia E. Kavraki},
    title = {{An Incremental Constraint-Based Framework for Task and Motion Planning}},
    number = {10},
    journal = {International Journal of Robotics Research, Special Issue on the 2016 Robotics: Science and Systems Conference},
    volume = {37},
    pages = {1134--1151},
    year = {2018}
}

@article{karaman2011IJRR,
  title={Sampling-based algorithms for optimal motion planning},
  author={Karaman, Sertac and Frazzoli, Emilio},
  journal={The International Journal of Robotics Research},
  volume={30},
  number={7},
  pages={846--894},
  year={2011},
  publisher={Sage Publications Sage UK: London, England}
}

@article{lagriffoul2014IJRR,
  title={Efficiently combining task and motion planning using geometric constraints},
  author={Lagriffoul, Fabien and Dimitrov, Dimitar and Bidot, Julien and Saffiotti, Alessandro and Karlsson, Lars},
  journal={The International Journal of Robotics Research},
  volume={33},
  number={14},
  pages={1726--1747},
  year={2014},
  publisher={SAGE Publications Sage UK: London, England}
}

@inproceedings{basch2001ICRA,
  title={Disconnection proofs for motion planning},
  author={Basch, Julien and Guibas, Leonidas J and Hsu, David and Nguyen, An Thai},
  booktitle={Proceedings 2001 ICRA. IEEE International Conference on Robotics and Automation (ICRA)},
  volume={2},
  pages={1765--1772},
  year={2001},
  organization={IEEE}
}

@article{thomas2021RAS,
title = {{MPTP: Motion-planning-aware task planning for navigation in belief space}},
journal = {Robotics and Autonomous Systems},
volume = {141},
pages = {103786},
year = {2021},
issn = {0921-8890},
doi = {https://doi.org/10.1016/j.robot.2021.103786},
url = {https://www.sciencedirect.com/science/article/pii/S0921889021000713},
author = {Antony Thomas and Fulvio Mastrogiovanni and Marco Baglietto}
}

@inproceedings{thomas2022IAS,
author={Thomas, Antony
and Mastrogiovanni, Fulvio},
title={{Minimum Displacement Motion Planning for Movable Obstacles}},
bookTitle={Intelligent Autonomous Systems 17},
year={2023},
publisher={Springer Nature Switzerland},
address={Cham},
pages={155--166},
isbn={978-3-031-22216-0}
}

@inproceedings{thomas2023IAS,
author={Thomas, Antony
and Mastrogiovanni, Fulvio and Baglietto, Marco},
title={{Revisiting the Minimum Constraint Removal Problem in Mobile Robotics}},
booktitle={Intelligent Autonomous Systems 18},
year={2024},
publisher={Springer Nature Switzerland},
address={Cham},
pages={31--41},
isbn={978-3-031-44851-5}
}

@inproceedings{hauser2013RSS,
  title={Minimum constraint displacement motion planning},
  author={Hauser, Kris},
booktitle={Proceedings of Robotics: Science and Systems IX},
month = {June},
address = {Berlin, Germany},
  doi       = {10.15607/RSS.2013.IX.017},
  year={2013}
}

@article{hauser2014IJRR,
  title={The minimum constraint removal problem with three robotics applications},
  author={Hauser, Kris},
  journal={The International Journal of Robotics Research},
  volume={33},
  number={1},
  pages={5--17},
  year={2014},
  publisher={SAGE Publications Sage UK: London, England}
}

@inproceedings{krontiris2015RSS,
  title={{Dealing with Difficult Instances of Object Rearrangement}},
  author={Krontiris, Athanasios and Bekris, Kostas E},
  booktitle={Proceedings of Robotics: Science and Systems XI},
  address  = {Rome, Italy}, 
month = {July},
  doi       = {10.15607/RSS.2015.XI.045},
  year={2015}
}

@article{stilman2005IJHR,
  title={Navigation among movable obstacles: Real-time reasoning in complex environments},
  author={Stilman, Mike and Kuffner, James J},
  journal={International Journal of Humanoid Robotics},
  volume={2},
  number={04},
  pages={479--503},
  year={2005},
  publisher={World Scientific}
}

@inproceedings{stilman2007ICRA,
  title={Manipulation planning among movable obstacles},
  author={Stilman, Mike and Schamburek, Jan-Ullrich and Kuffner, James and Asfour, Tamim},
  booktitle={Proceedings 2007 IEEE international conference on robotics and automation},
  pages={3327--3332},
  year={2007},
  organization={IEEE}
}

@inproceedings{dogar2011RSS,
 author={Dogar, Mehmet and Srinivasa, Siddhartha},
  title={A framework for push-grasping in clutter},
   booktitle={Proceedings of Robotics: Science and Systems VII},
month = {June},
address = {Los Angeles, CA, USA},
editor = {Hugh Durrant-Whyte, Nick Roy and Pieter Abbeel},
publisher = {MIT Press},
doi = {10.15607/RSS.2011.VII.009},
   year={2011}
}

@inproceedings{karami2021AIIA,
author={Karami, Hossein and Thomas, Antony and Mastrogiovanni, Fulvio},
title={{Task Allocation for Multi-robot Task and Motion Planning: A Case for Object Picking in Cluttered Workspaces}},
booktitle={{AIxIA 2021 -- Advances in Artificial Intelligence}},
year={2022},
publisher={Springer International Publishing},
address={Cham},
pages={3--17},
isbn={978-3-031-08421-8}
}

@inproceedings{thomas2023ICRA,
  author={Thomas, Antony and Ferro, Giulio and Mastrogiovanni, Fulvio and Robba, Michela},
  booktitle={2023 IEEE International Conference on Robotics and Automation (ICRA)}, 
  title={Computational Tradeoff in Minimum Obstacle Displacement Planning for Robot Navigation}, 
  year={2023},
  volume={},
  number={},
  pages={3635-3641},
  doi={10.1109/ICRA48891.2023.10161372}}

@article{li2023IJRR,
  title={A sampling and learning framework to prove motion planning infeasibility},
  author={Li, Sihui and Dantam, Neil T},
  journal={The International Journal of Robotics Research},
  volume={42},
  number={10},
  pages={938--956},
  year={2023},
  publisher={SAGE Publications Sage UK: London, England}
}

@inproceedings{zhang2007IROS,
  title={A hybrid approach for complete motion planning},
  author={Zhang, Liangjun and Kim, Young J and Manocha, Dinesh},
  booktitle={2007 IEEE/RSJ International Conference on Intelligent Robots and Systems},
  pages={7--14},
  year={2007},
  organization={IEEE}
}

@article{zhang2008IJRR,
  title={Efficient cell labelling and path non-existence computation using C-obstacle query},
  author={Zhang, Liangjun and Kim, Young J and Manocha, Dinesh},
  journal={The International Journal of Robotics Research},
  volume={27},
  number={11-12},
  pages={1246--1257},
  year={2008},
  publisher={SAGE Publications Sage UK: London, England}
}

@inproceedings{mccarthy2012ICRA,
  title={Proving path non-existence using sampling and alpha shapes},
  author={McCarthy, Zoe and Bretl, Timothy and Hutchinson, Seth},
  booktitle={2012 IEEE international conference on robotics and automation},
  pages={2563--2569},
  year={2012},
  organization={IEEE}
}

@article{varava2021IJRR,
  title={Free space of rigid objects: Caging, path non-existence, and narrow passage detection},
  author={Varava, Anastasiia and Carvalho, J Frederico and Kragic, Danica and Pokorny, Florian T},
  journal={The International Journal of Robotics Research},
  volume={40},
  number={10-11},
  pages={1049--1067},
  year={2021},
  publisher={SAGE Publications Sage UK: London, England}
}

@inproceedings{sung2023RSS, 
    author    = {Yoonchang Sung AND Peter Stone}, 
    title     = {{Motion Planning (In)feasibility Detection using a Prior Roadmap via Path and Cut Search}}, 
    booktitle = {Proceedings of Robotics: Science and Systems}, 
    year      = {2023}, 
    address   = {Daegu, Republic of Korea}, 
    month     = {July}, 
    doi       = {10.15607/RSS.2023.XIX.060} 
}

@article{wells2019RAL,
  title={Learning feasibility for task and motion planning in tabletop environments},
  author={Wells, Andrew M and Dantam, Neil T and Shrivastava, Anshumali and Kavraki, Lydia E},
  journal={IEEE robotics and automation letters},
  volume={4},
  number={2},
  pages={1255--1262},
  year={2019},
  publisher={IEEE}
}

@article{driess2021IJRR,
  title={Learning to solve sequential physical reasoning problems from a scene image},
  author={Driess, Danny and Ha, Jung-Su and Toussaint, Marc},
  journal={The International Journal of Robotics Research},
  volume={40},
  number={12-14},
  pages={1435--1466},
  year={2021},
  publisher={SAGE Publications Sage UK: London, England}
}

@inproceedings{li2020IROS,
  title={Towards general infeasibility proofs in motion planning},
  author={Li, Sihui and Dantam, Neil T},
  booktitle={2020 IEEE/RSJ International Conference on Intelligent Robots and Systems (IROS)},
  pages={6704--6710},
  year={2020},
  organization={IEEE}
}

@article{kavraki1995TRO,
  title={Computation of configuration-space obstacles using the fast Fourier transform},
  author={Kavraki, Lydia E},
  journal={IEEE Transactions on Robotics and Automation},
  volume={11},
  number={3},
  pages={408--413},
  year={1995},
  publisher={IEEE}
}

@inproceedings{curto1997CIRA,
  title={Mathematical formalism for the fast evaluation of the configuration space},
  author={Curto, Bel{\'e}n and Moreno, Vidal},
  booktitle={Proceedings 1997 IEEE International Symposium on Computational Intelligence in Robotics and Automation CIRA'97.'Towards New Computational Principles for Robotics and Automation'},
  pages={194--199},
  year={1997},
  organization={IEEE}
}

@article{mccoid2022TMS,
  title={A provably robust algorithm for triangle-triangle intersections in floating-point arithmetic},
  author={Mccoid, Conor and Gander, Martin J},
  journal={ACM Transactions on Mathematical Software (TOMS)},
  volume={48},
  number={2},
  pages={1--30},
  year={2022},
  publisher={ACM New York, NY}
}

@inproceedings{muguiraIturralde2023ICRA,
  author={Muguira-Iturralde, Jose and Curtis, Aidan and Du, Yilun and Kaelbling, Leslie Pack and Lozano-Pérez, Tomás},
  booktitle={2023 IEEE International Conference on Robotics and Automation (ICRA)}, 
  title={{Visibility-Aware Navigation Among Movable Obstacles}}, 
  year={2023},
  pages={10083-10089},
  doi={10.1109/ICRA48891.2023.10160865}
  }

@article{li2023RAL,
  title={{Scaling infeasibility proofs via concurrent, codimension-one, locally-updated coxeter triangulation}},
  author={Li, Sihui and Dantam, Neil T},
  journal={IEEE Robotics and Automation Letters},
  volume={8},
  number={12},
  pages={8303--8310},
  year={2023},
  publisher={IEEE}
}

@inproceedings{orthey2018IROS,
  title={Quotient-space motion planning},
  author={Orthey, Andreas and Escande, Adrien and Yoshida, Eiichi},
  booktitle={2018 IEEE/RSJ International Conference on Intelligent Robots and Systems (IROS)},
  pages={8089--8096},
  year={2018},
  organization={IEEE}
}

@inproceedings{orthey2019ISRR,
  title={Rapidly-exploring quotient-space trees: Motion planning using sequential simplifications},
  author={Orthey, Andreas and Toussaint, Marc},
  booktitle={The International Symposium of Robotics Research},
  pages={52--68},
  year={2019},
  organization={Springer}
}

@inproceedings{montaut2022RSS, 
    author    = {Louis Montaut and {Quentin Le} Lidec and Vladimír Petrík and Josef Sivic and Justin Carpentier}, 
    title     = {{Collision Detection Accelerated: An Optimization Perspective}}, 
    booktitle = {Proceedings of Robotics: Science and Systems}, 
    year      = {2022}, 
    address   = {New York City, NY, USA}, 
    month     = {June}, 
    doi       = {10.15607/RSS.2022.XVIII.039} 
}

@article{das2020TRO,
  title={Learning-based proxy collision detection for robot motion planning applications},
  author={Das, Nikhil and Yip, Michael},
  journal={IEEE Transactions on Robotics},
  volume={36},
  number={4},
  pages={1096--1114},
  year={2020},
  publisher={IEEE}
}
\end{document}